\documentclass[10pt,journal,table,hidelinks]{IEEEtran}
\usepackage{hyperref}
\usepackage{amsmath}
\usepackage{graphicx}
\usepackage[caption=false]{subfig}
\usepackage{color}
\usepackage{booktabs}
\usepackage{multirow}
\usepackage{array}
\usepackage{framed} 
\usepackage{tabularx}
\newlength{\mytextsize}
\usepackage[table]{xcolor} 

\newcommand{\emoji}[1]{\includegraphics[height=\mytextsize]{emoji/#1.pdf}}
\newcommand{\emojipng}[1]{\includegraphics[height=\mytextsize]{emoji/#1.png}}

\ifCLASSOPTIONcompsoc
  \usepackage[nocompress]{cite}
\else
  \usepackage{cite}
\fi
\ifCLASSINFOpdf
\else
\fi

\newcommand{\eg}{e.g. }

\newcommand{\etal}{\mbox{\emph{et al.\ }}}

\begin{document}
%
\title{The New Modality:\\ Emoji Challenges in Prediction, \\Anticipation, and Retrieval}
%
%
%
%

\author{Spencer~Cappallo, Stacey~Svetlichnaya, Pierre~Garrigues, Thomas~Mensink, Cees~G.M.~Snoek}
\markboth{}
{Shell \MakeLowercase{\textit{et al.}}: Bare Demo of IEEEtran.cls for Computer Society Journals}
%



\IEEEtitleabstractindextext{%
\begin{abstract}
Over the past decade, emoji have emerged as a new and widespread form of digital communication, spanning diverse social networks and spoken languages. We propose to treat these ideograms as a new modality in their own right, distinct in their semantic structure from both the text in which they are often embedded as well as the images which they resemble. As a new modality, emoji present rich novel possibilities for representation and interaction. In this paper, we explore the challenges that arise naturally from considering the emoji modality through the lens of multimedia research. Specifically, the ways in which emoji can be related to other common modalities such as text and images. To do so, we first present a large scale dataset of real-world emoji usage collected from Twitter. This dataset contains examples of both text-emoji and image-emoji relationships. We present baseline results on the challenge of predicting emoji from both text and images, using state-of-the-art neural networks. Further, we offer a first consideration into the problem of how to account for new, unseen emoji -- a relevant issue as the emoji vocabulary continues to expand on a yearly basis. Finally, we present results for multimedia retrieval using emoji as queries. 

\end{abstract}

}

\maketitle

\IEEEdisplaynontitleabstractindextext

%
\IEEEpeerreviewmaketitle

\section{Introduction}\label{sec:introduction}

%
%
%
%

\IEEEPARstart{E}{moji}, small ideograms depicting objects, people, and scenes, have exploded in popularity. They are now available on all major mobile phone platforms and social media websites, as well as many other places. According to the \emph{Oxford English Dictionary}, the term \emph{emoji} is a Japanese coinage meaning `pictogram', created by combining \emph{e} (picture) with \emph{moji} (letter or character). Emoji as we know them were first introduced as a set of 176 pictogram available to users to Japanese mobile phones. The available range of ideograms has expanded greatly over the previous years, with 1,144 single emoji characters defined in Unicode 10.0 and many more defined through combinations of two or more emoji characters. In this paper, we approach emoji as a modality related to, but not contained within, text and images. We investigate the properties and challenges of relating these modalities to emoji, as well as the multimedia retrieval opportunities that emoji present. 


The identification and benchmarking of novel modalities has a rich history in the multimedia community. 
When new modalities are identified, it is important to make first attempts to understand their relationship with already established information channels. 
One way in which to do this is to explore the cross-modal relationships between the modality and other modalities. 
When Lee \etal \cite{lee2010predicting} identified nonverbal head nods as an information-rich and overlooked modality, they sought to provide understanding through prediction of them based on semantic understanding of the accompanying conversation transcript. 
Like emoji, new modalities are sometimes the result of a newly developed technology, as with 3D models~\cite{gao2014view} or the growth of microblogging \cite{aiello2013sensing}. 
Though ideograms are ancient, emoji is a modern technological evolution of that ancient idea. 
The march of technology sometimes facilitates new looks at old problems, such as the use of infrared imagery for facial recognition instead of natural images \cite{wang2010natural}. 
Often, the presentation of new tasks as research challenges can accelerate research progress, as it did with acoustic scenes~\cite{stowell2015detection} and video concepts~\cite{snoek2006challenge}. 
We look to this history of multimedia challenge problems and identify emoji as an emerging modality worthy of a similar treatment. 
To facilitate further research on emoji, we propose three emoji challenge problems and present state-of-the-art neural network baselines for them, as well as a dataset for evaluation.

Despite their prevalence, research into emoji remains limited. The majority of prior research concerning emoji has focused on descriptive analysis, such as identifying how patterns of emoji usage shift among different demographics~\cite{barbieri2016cosmopolitan,chen2017through}, or have used them as a signal to indicate the emotional affect of accompanying media~\cite{rathan2017consumer,guthierlanguage}. The focus on sentiment is likely a result of there being a number of ``face emoji" (\eg \emojipng{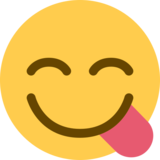}) which are designed to exhibit a particular emotion or reaction. These face emoji are by far the most visible emoji and among the most widely used~\cite{pohl2017beyond}, but the focus on them ignores the hundreds of other emoji which are worthy of study in their own right. Beyond these face emoji, the full set of emoji also contains a wide range of other objects, such as foods (\emojipng{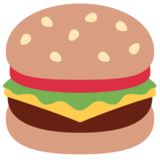}), signs (\emojipng{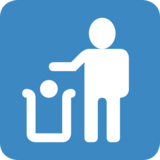}), and scenes (\emojipng{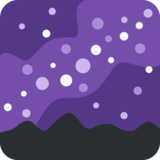}) which may lack a strong sentimental signal~\cite{novak2015sentiment}. Focusing solely on the emotion-laden subset of emoji ignores the information conveyed and possibilities presented by the many other ideograms available.

\begin{figure*}
\centering
\begin{tabular*}{0.95\textwidth}{@{\extracolsep{\stretch{1}}}*{6}{c}@{} }
(00:08.33) & (00:16.67) & (00:25.00) & (00:33.33) & (00:41.67) & Entire Video \\
\includegraphics[width=0.12\textwidth]{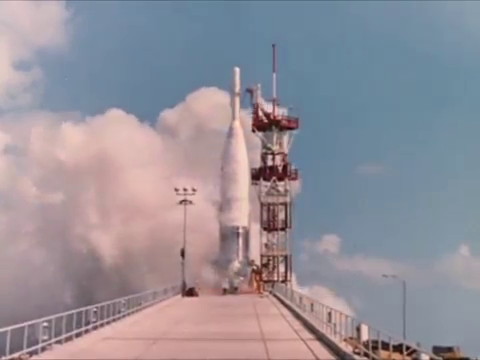} & 
\includegraphics[width=0.12\textwidth]{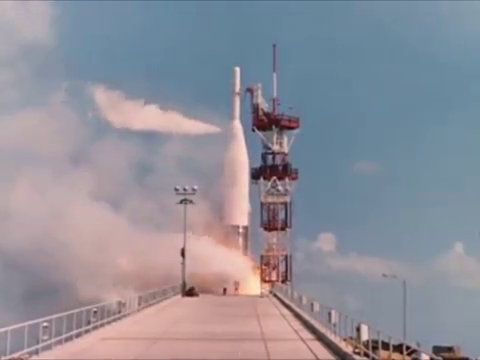} &
\includegraphics[width=0.12\textwidth]{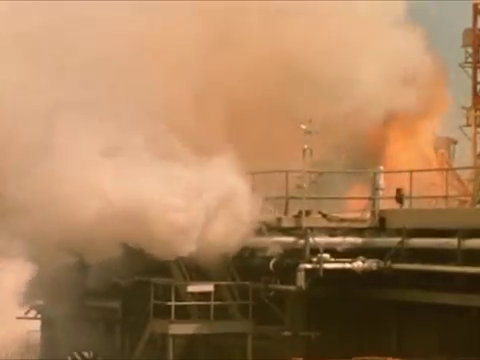} &
\includegraphics[width=0.12\textwidth]{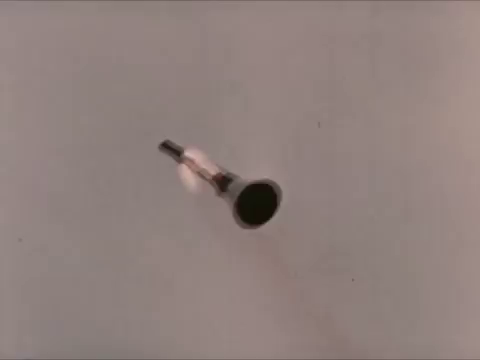} &
\includegraphics[width=0.12\textwidth]{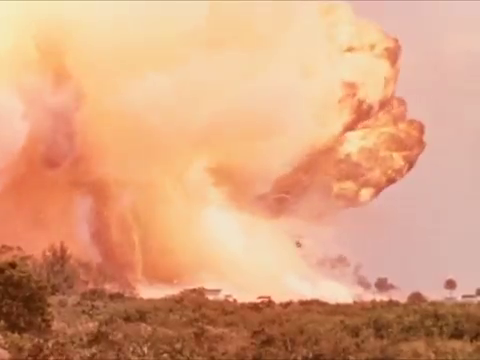} & 
\includegraphics[width=0.12\textwidth]{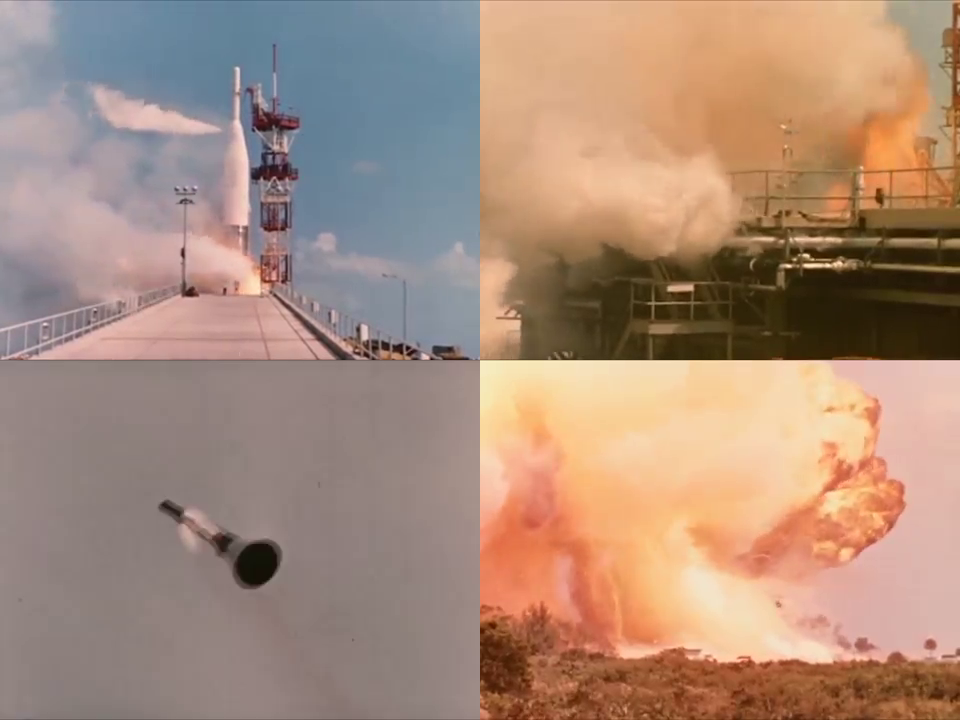} \\
\includegraphics[height=1\mytextsize]{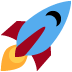}
\includegraphics[height=1\mytextsize]{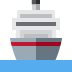}
\includegraphics[height=1\mytextsize]{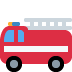}
\includegraphics[height=1\mytextsize]{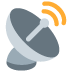}
\includegraphics[height=1\mytextsize]{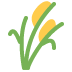} &
\includegraphics[height=1\mytextsize]{kept_graphics/rocket.png}
\includegraphics[height=1\mytextsize]{kept_graphics/ship.png}
\includegraphics[height=1\mytextsize]{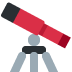}
\includegraphics[height=1\mytextsize]{kept_graphics/fire_engine.png}
\includegraphics[height=1\mytextsize]{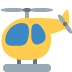} &
\includegraphics[height=1\mytextsize]{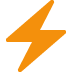}
\includegraphics[height=1\mytextsize]{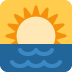}
\includegraphics[height=1\mytextsize]{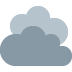}
\includegraphics[height=1\mytextsize]{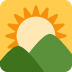}
\includegraphics[height=1\mytextsize]{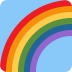} &
\includegraphics[height=1\mytextsize]{kept_graphics/lightning.png}
\includegraphics[height=1\mytextsize]{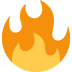}
\includegraphics[height=1\mytextsize]{kept_graphics/cloud.png}
\includegraphics[height=1\mytextsize]{kept_graphics/sunrise.png}
\includegraphics[height=1\mytextsize]{kept_graphics/rocket.png} &
\includegraphics[height=1\mytextsize]{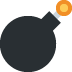}
\includegraphics[height=1\mytextsize]{kept_graphics/rocket.png}
\includegraphics[height=1\mytextsize]{kept_graphics/fire_engine.png}
\includegraphics[height=1\mytextsize]{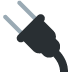}
\includegraphics[height=1\mytextsize]{kept_graphics/fire.png} 
&
\includegraphics[height=1\mytextsize]{kept_graphics/bomb.png}
\includegraphics[height=1\mytextsize]{kept_graphics/rocket.png}
\includegraphics[height=1\mytextsize]{kept_graphics/fire_engine.png}
\includegraphics[height=1\mytextsize]{kept_graphics/fire.png}
\includegraphics[height=1\mytextsize]{kept_graphics/satellite_dish.png}
\\
\end{tabular*}
\\\vspace{0.5em}
\textbf{A.}

\vspace{1em}
\begin{tabular}{ m{0.025\textwidth} m{0.42\textwidth} m{0.025\textwidth} m{0.42\textwidth}}
\includegraphics[height=1\mytextsize]{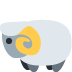}& \includegraphics[width=0.42\textwidth]{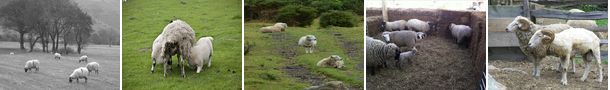} &\includegraphics[height=1\mytextsize]{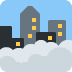}& \includegraphics[width=0.42\textwidth]{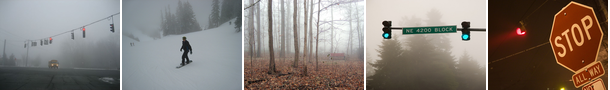}\\
\includegraphics[height=1\mytextsize]{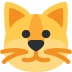} & \includegraphics[width=0.42\textwidth]{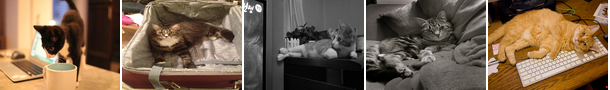} & \includegraphics[height=1\mytextsize]{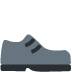} & \includegraphics[width=0.42\textwidth]{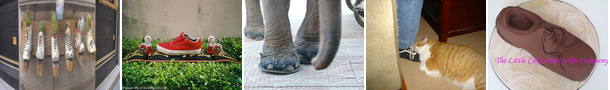}\\
\includegraphics[height=2\mytextsize]{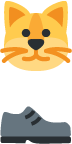} & \includegraphics[width=0.42\textwidth]{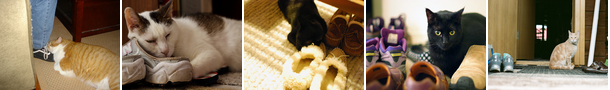}& & \\
\end{tabular}\\\vspace{0.5em}
\textbf{B.}
\caption{Emoji prediction used for video summarization and query-by-emoji, adapted from our previous work~\cite{cappallo2015emoji}. \textbf{A}. The emoji summarization of the entire video presents a more complete representation of the video's contents than a single screenshot might. \textbf{B}. Emoji can be used as a language-agnostic query language for media retrieval tasks. Here, emoji are used to retrieve photos from the MSCOCO dataset. Despite their limited vocabularity, emoji can be combined to compose more nuanced queries, such as \emph{shoe}$+$\emph{cat}. This results in a surprisingly flexible modality for both content description and retrieval.}
\label{fig:catandshoe}
\end{figure*}

In this work, we approach emoji as an information-rich modality in their own right. Though emoji are commonly embedded in text, we view them as distinct from text. 
Their visual nature allows for emoji to add richness of meaning and variety of semantics that is unavailable in pure text. 
When embedded in text, emoji sometimes simply replace a word, but more often they provide new information which was not contained in the text alone~\cite{ai2017untangling,na2017varying}. 
Emoji can be used as a supplemental modality to clarify the intended sense of an ambiguous message~\cite{riordan2017communicative}, attach sentiment to a message~\cite{shiha2017effects}, or subvert the original meaning of the text entirely in ways a word could not~\cite{donato2017investigating,njenga2017social}. 
Emoji carry meaning on their own, and possess compositionality allowing for more nuanced semantics through multi-emoji phrases~\cite{lopez2017did}. 
Many emoji are used in cases where the particular symbol resembles something else entirely, acting as a kind of visual pun. 
These qualities, along with a cross-language similarity of semantics~\cite{barbieri2016cosmopolitan}, suggest that emoji, despite being unicode characters, are distinct from their frequent textual bedfellows. 

Though emoji are represented by small pictures, they are distinct from standard images. As a form of symbology, the specifics of the individual representation is often incidental to the underlying meaning of the ideogram -- this is unlike images where the particulars of a given image are often more crucial than what it is representing generally (\emph{i.e.}, it is a photo of \emph{your} dog, not just a photo representing the semantic notion of `dog'). This difference is further substantiated by the fact that emoji exist as nothing more than unicode characters. As characters, the details of their illustrations are left up to the platform supporting them, and significant variation for a single emoji can exist between platforms~\cite{miller2016blissfully,tigwell2016oh}. For these reasons, their behaviour and meaning is substantially different from that of images. Figure \ref{fig:catandshoe} gives examples of video summary using emoji and query-by-emoji, which nicely demonstrate the way in which emoji as ideograms are related to but different from natural imagery. 

Having established the view that emoji constitute a distinct modality from text or images, this paper seeks to explore the ramifications of this viewpoint through the lens of multimedia retrieval challenges. As a modality, we focus on the relationship between emoji and two other modalities, namely text and images.
%
%
This work makes the following contributions.
\begin{itemize}
\item We propose and support the treatment of emoji as a modality distinct from either text or images.
\item We present a large scale dataset composed of real-world emoji usage, containing both textual and text+image examples. We consider a wide range of over 1000 emoji, including the often overlooked long tail of emoji. This data set as well as the training splits will be available for future researchers.
\item We propose three challenge tasks for relating emoji to text and images, and present state-of-the-art baseline results on these. Namely, the tasks are emoji prediction from text and/or images, prediction of unanticipated emoji using their unicode description, and lastly multimedia retrieval using emoji as queries.
\end{itemize}

In the following section we give an overview of previous work on emoji. In Section \ref{sec:newmodality} we present our dataset, and propose three challenge tasks presented by the emoji modality. In Sections \ref{sec:supervisedprediction}, \ref{sec:zeroshotemoji}, and \ref{sec:querybyemoji} we present baseline results for each of these challenge tasks using state-of-the-art deep learning approaches. In Section \ref{sec:conclusion}, we conclude.

\section{Related Work}
\label{sec:relatedwork}

\subsection{Emoji}
Previous work on emoji in the scientific community has focused on using them as a source of sentiment annotation, or on descriptive analysis of emoji usage.

\subsubsection{Emoji for Sentiment}
Much of prior work has viewed emoji primarily as an indicator of sentiment. This is done either explicitly, through the direct consideration of sentiment, or implicitly, through the consideration of only popular emoji. The most popular emoji are disproportionately composed of sentiment-laden emoji. 
Face emojis, thumbs-up, and hearts have high incidence, while less emotional emoji such as symbols, objects, and flags, have much lower incidence. The result is that any work which considers only the most popular emoji may have an inherent bias toward heavy sentiment emoji.

Several works look at the effect that including emoji can have on the perception of accompanying text. Some find that the inclusion of emoji increases the perceived level of sentiment attached to a message  \cite{shiha2017effects,na2017varying,novak2015sentiment}. Similarly, the work from \cite{rodrigues2017frown} finds that emoji correlate to a more positive perception for messages in a dating app than messages that don't contain emoji. These works demonstrate that emoji can be a useful supplementary signal for sentiment within text messages, but these works focus primarily on face emoji designed specifically for the communication of emotion. In contrast, \cite{riordan2017communicative} investigates the affect of non-face emoji. They found that even non-face emoji can increase perceived emotion, and also can improve clarity of text that is otherwise ambiguous. Some text phrases are ambiguous when considered alone, but the inclusion of another modality (emoji) can help readers to pin-point the intended sense (\eg ``I took the shot" vs ``I took the shot \emojipng{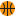}").

A notable work of sentiment analysis of emoji is \cite{novak2015sentiment}, which annotated a collection of tweets with sentiment and presented sentiment rankings for 751 emoji (the most frequent in their data). Their work demonstrated that while some emoji have very strong positive sentiment scores, others were very neutral, being rarely associated with strong positive or negative sentiment. Similarly, they observed that some emoji are used frequently to denote both strong positive and negative sentiment. These observations suggest that treating emoji as merely a straightforward signal of sentiment is misguided, and that there's a more nuanced richness and variety to emoji meaning.

Lastly, some works consider emoji, particularly face emoji, as a pure sentiment signal. The approach by \cite{rathan2017consumer} incorporates emoji as an input source for evaluating the sentiment of social media messages mentioning particular brands. Going a step further, \cite{guthierlanguage} assumes emoji to be a reliable ground truth for sentiment. They construct a dataset for sentiment prediction and use a set of emoji to automatically annotate the dataset. Given the broad ambiguity of usage and the sentiment gap between emoji and text explored in other works, such an approach may yield noisy annotation.

\subsubsection{Analysis of Emoji Usage}
Numerous works have helped to glean insight into the properties and trends of real-world emoji usage. Several have looked at the manner in which emoji usage varies between different countries and cultures  \cite{ljubevsic2016global,barbieri2016cosmopolitan,lu2016learning}. Meanwhile \cite{chen2017through} analyzes differences in emoji usage patterns between genders. While there are differences between how specific communities may use emoji, the data makes clear that emoji usage is on the rise globally\cite{ljubevsic2016global,zhou2017goodbye}. This further supports our viewpoint that emoji are their own modality, as they are not tied to any one particular culture or language and share semantic commonalities which are orthogonal to the community that uses them.

Several works look at the problem of ambiguity in the perceived meaning of emoji \cite{miller2016blissfully,tigwell2016oh,miller2017understanding}. In general, they find a degree of ambiguity with emoji, and that the choice of illustration used by a particular platform (\eg iOS or Android) can increase this confusion. Notably, \cite{miller2017understanding} observes that the inclusion of an additional input modality (in the form of textual context) improves the distinctiveness of meaning substantially. This observation is well in line with what has been known in the multimedia community for years: that a multi-modal approach can improve prediction. Ambiguity between the message intent from the author of an emoji-containing message and its interpretation by readers has also been investigated \cite{berengueres2017sentiment}. The ambiguity and breadth of possible meaning for a given emoji helps to make emoji a challenging modality for algorithmic understanding, worthy of pursuing and with a high ceiling for perfection.

The relationship among emoji themselves has been studied in \cite{pohl2017beyond,barbieri2016does,wijeratne2017semantics}. The work of \cite{pohl2017beyond} gives a thorough analysis of emoji usage, and proposes a model for analyzing the relatedness of pairs of emoji. Similarly, \cite{barbieri2016does} looks at the problem of trying to identify text tokens which are most closely related to a given emoji. The authors do this by learning a shared embedding space using a skip-gram model \cite{mikolov2013efficient}, and identifying those text tokens closest to the emoji within this mutual semantic space. While both \cite{pohl2017beyond} and \cite{barbieri2016does} learn models that could be applied to emoji prediction, they both focus instead on descriptive analysis of emoji usage.

Along similar lines, there has been some recent work on identifying the different ways in which emoji can be used in combination with text. \cite{donato2017investigating,na2017varying,ai2017untangling} use emoji either as a straightforward replacement for text, or as a supplementary contribution which alters or enhances the meaning of the text. The work of \cite{donato2017investigating} constructs a dataset of 4100 tweets that have been annotated to indicate whether the emoji contain redundant information (already contained in the text) or not. Among their collection of annotated tweets, they found that the non-redundant class was the largest class of emoji. This result supports our proposition that emoji are distinct from, though entwined with, any text that accompanies them. 

While works such as \cite{barbieri2016does,ai2017untangling,pohl2017beyond} tackle the problem of understanding emoji usage through building models on top of real world usage data, there has also been work on trying to build an emoji understanding in a more hand-crafted fashion. For example, \cite{wijeratne2016emojinet} acquires a structured understanding of emoji usage through combining several user-defined databases of emoji meaning. Their later work then uses this data to learn a model for sentiment analysis which performs comparably to models trained directly on real world usage data \cite{wijeratne2017semantics}. This kind of structured, pre-defined understanding of emoji is similar to the no-example approach explored in our previous work \cite{cappallo2015emoji} and further explored in this work. This work, however, targets emoji as a rich, informative modality rather than only a means to perform sentiment analysis.

\cite{lopez2017did} is an early investigation into the compositionality of emoji. They find that emoji can be combined to create new composed meanings, a finding which lends support to the notion of composing queries from multiple emojis that is discussed in this work.

\begin{figure*}
\centering
\includegraphics[width=0.7\textwidth]{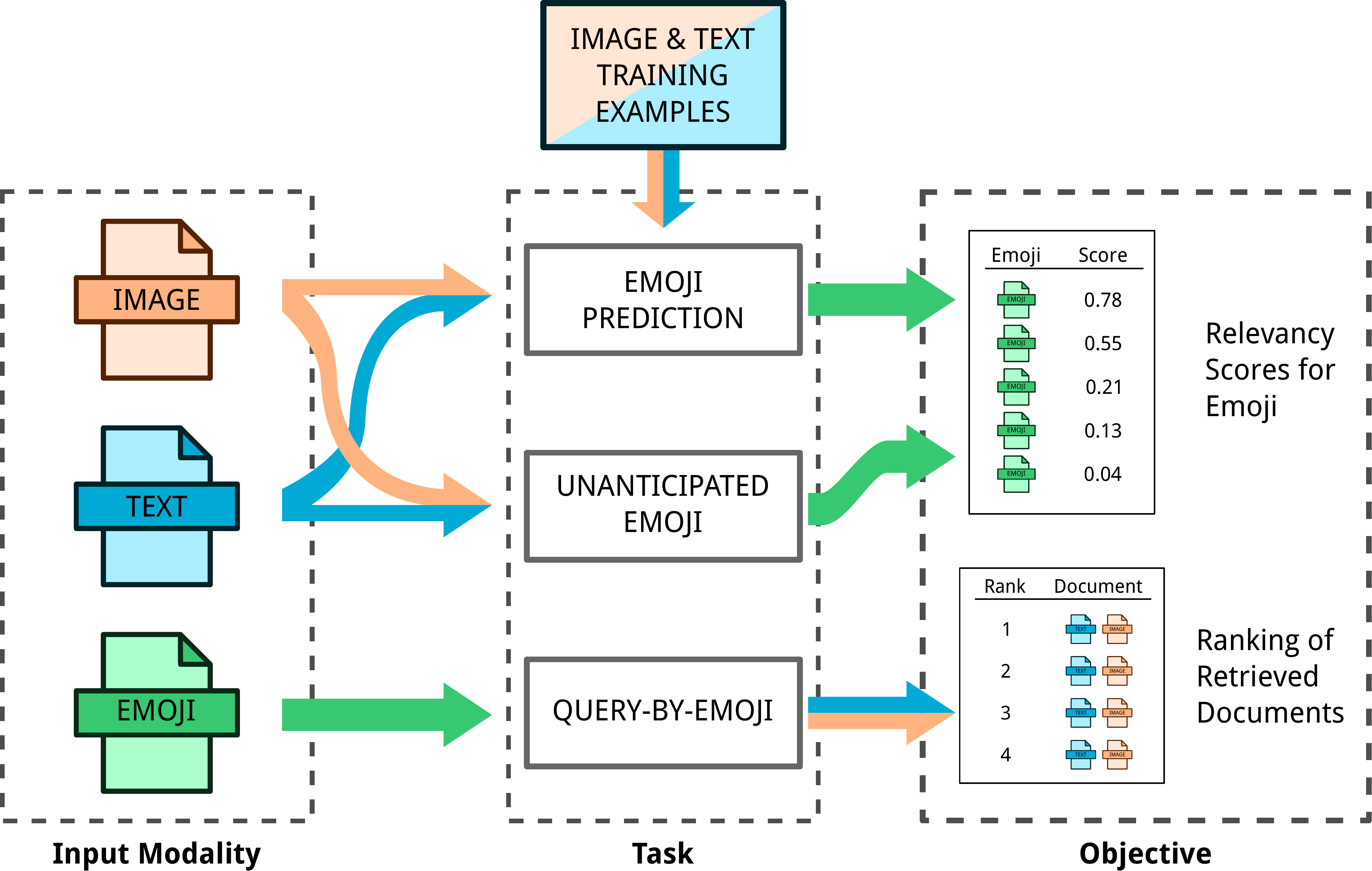}
\caption{Overview of our three proposed tasks. Emoji Prediction and Unanticipated Emoji both seek to score emoji based on other input modalities. Their difference is that Emoji Prediction has the benefit of emoji-annotated training examples to learn from, while Unanticipated Emoji simulates the setting of newly released emoji where there is no training data available. Query-by-Emoji seeks to retrieve relevant multi-modal documents using queries composed with emoji.}
\label{fig:taskoverview}
\end{figure*}

Much of the analysis of these works support our philosophy of treating emoji as a modality in their own right.
In contrast to these works and to complement them, rather than trying to provide descriptive analysis of emoji usage, we focus on how the emoji can be used with and related to other modalities.


\subsubsection{Cross-modal Emoji Prediction}

A few recent works have investigated the problem of emoji prediction, which is closer to our position of emoji-as-modality. 

Our previous work was the first to look at the problem of emoji prediction~\cite{cappallo2015emoji}, and approached from a zero-shot perspective due to a lack of established dataset. Following on from the work, a query-by-emoji video search engine was also proposed~\cite{cappallo2015query}. These works reported quantitative results only on related tasks in other modalities, and presented only qualitative results for the emoji modality. We instead present results on a large scale, real-world emoji dataset, with proposed tasks and state-of-the-art supervised baselines.

Felbo et al.~\cite{felbo2017using} learn a model to predict emoji based on input text. Rather than using the model directly for the task of emoji prediction, they use this model as a form of pre-training for learning a sentiment prediction network. Additionally, their emoji model is intentionally limited to 64 emoji chosen for having a high degree of sentiment. Our aim is to treat emoji as an end goal rather than an intermediary, and to consider the full breadth of emoji available including rare emoji or emoji with little or no sentiment attached to them.

Barbieri et al.~\cite{barbieri2017emojis} looked at the problem of emoji prediction based on an input text. Their setting is most similar to the one considered in this paper. However, they focus strictly on text, while we consider also images. Further, Barbieri et al. restrict their labels to only the top 20 most frequent emoji within their dataset. Along similar lines, \cite{li2017joint} uses a convolutional network to predict 100 common emoji based on a corresponding text from weibo or another social media network. Both of these papers consider only the most common emoji. There are thousands of emoji, and the longtail of the available emoji are a valuable and difficult prediction task. We consider the full range of emoji present in our dataset, and look at the problems involved with tackling this longtail. We further distinguish our work by also considering the problem of newly introduced emoji, which is important as the set of available ideograms is growing every year.

El et al.~\cite{el2017face2emoji} is, to the best of our knowledge, the only previous work that considers supervised prediction of emoji from images. Their work looks at the problem of translating images of faces into corresponding face emojis. We take a broader approach both on the image and annotation sides, seeking to instead predict any sort of relevant emoji based on a wide variety of images.

\section{New Modality}
\label{sec:newmodality}


There is no guarantee that a simple explanation of what an emoji depicts will encompass its full semantic burden. 
Emoji are inherently representational, so by definition some overlap in semantics is expected, but that overlap may be incomplete in terms of real-world usage. 
For example, the emoji for \emph{cactus} \emoji{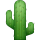} is not used only to represent a cactus, but is also widely used to signify a negative sentiment due to its resemblance to a certain hand gesture. 
This discrepancy between the intended semantics and the actual semantics leads us to propose learning the semantics directly from real-world usage in a large dataset collected from Twitter.

Motivated by our view that emoji constitute a separate modality, in this section we outline our methodological approach to establishing baseline analysis and results for the emoji modality. 
We begin by establishing three emoji challenge tasks, and subsequently propose a large dataset of real-world emoji usage as a testbed for exploring these challenges. 
We further propose evaluation criteria to quantify and compare performance on these challenges and dataset. An overview of how these three tasks differ in their objectives and the information available to them is provided in Figure \ref{fig:taskoverview}.

\subsection{Emoji Challenges}

\subsubsection{Emoji Prediction - How to predict emoji?}

There are thousands of emoji, and new ones are added every year. 
As they develop into an ever richer information signal, it is useful to understand how emoji are related to other modalities. 
The most straightforward way to go about this is to look at how well we can predict emoji given another, related input. 
%
Since emoji can be flexible in their usage, the question becomes: Given some input text and/or image, can we predict the relevant emoji that would accompany that input? 
This work seeks to present strong first baselines for the problem. 


We propose an Emoji Prediction challenge where the objective is to predict relevant emoji from alternative input modalities. Using real-world training examples correlating text and images to emoji annotations, models much seek to predict relevant emoji when presented with test examples.

\subsubsection{Emoji Anticipation - What to do about new emoji?}

A large real-world dataset provides the opportunity for learning how to use emoji in a natural way that reflects their true semantics. However, new emoji are added to the unicode specification every year, and will be deployed to users before their real world usage can be known. Any system that seeks to understand or suggest emoji to users should be prepared to deal with the challenge of new, previously unseen emoji.


In the Emoji Anticipation challenge, real world training data of emoji usage is no longer available. This simulates the situation when a new crop of emoji have been announced, but have not yet been deployed onto common platforms. Systems seeking to understand and predict these emoji must therefore exploit alternative knowledge sources. We present the problem as a zero-shot cross-modal problem, where we have only textual metadata regarding the emoji and must then try to determine its relevancy to images or text. This task shares some resemblance to that of zero-shot image classification \cite{conse, akata2016label} or zero example video retrieval \cite{jiang2014zero,chen2014event}. Generally, in zero-shot classification the model has a disjoint set of seen and unseen classes, and attempts to leverage the knowledge of seen classes as well as external information to classify the unseen classes. Our setting differs from this, as we test our model in a setting where it has seen no direct examples of the target modality whatsoever.


\begin{figure*}\centering
\begin{tabular}{c c c}
\includegraphics[width=0.25\textwidth]{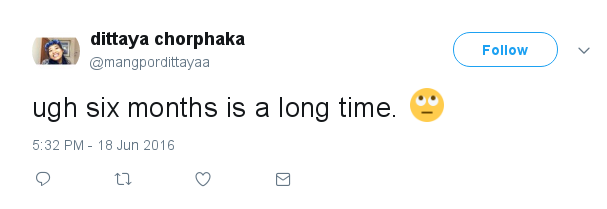} & \includegraphics[width=0.25\textwidth]{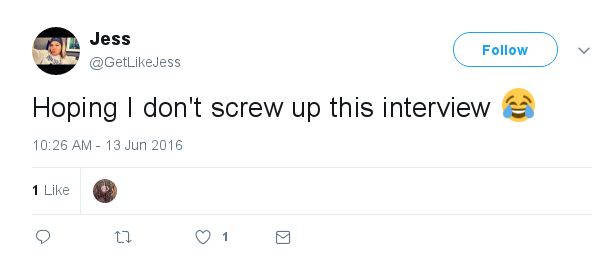} & \includegraphics[width=0.25\textwidth]{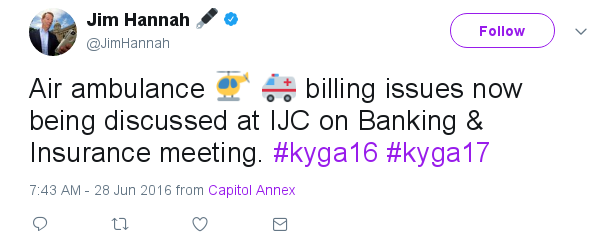}\\
A. & B. & C. \\
\includegraphics[width=0.25\textwidth]{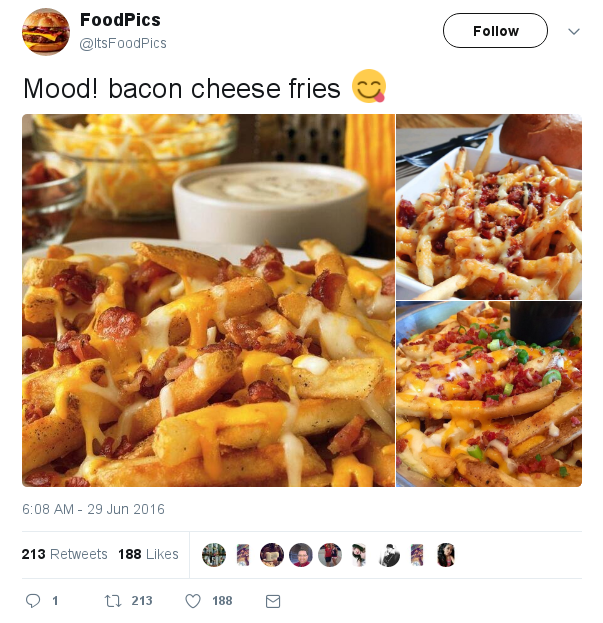} & \includegraphics[width=0.25\textwidth]{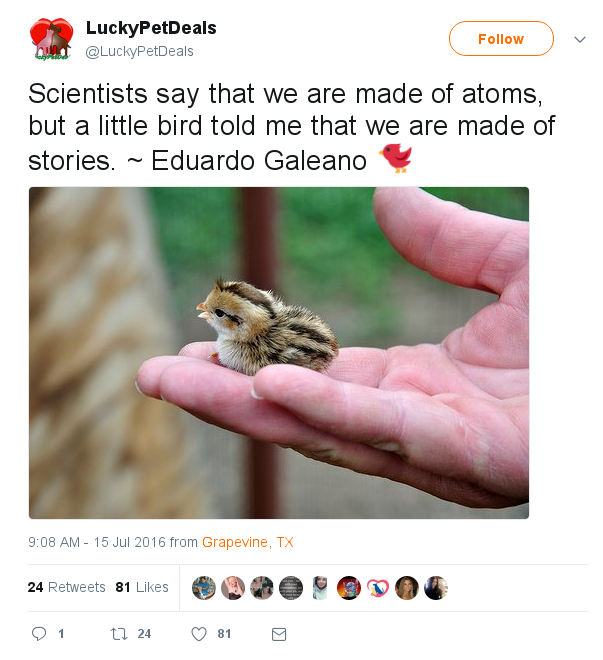} & \includegraphics[width=0.25\textwidth]{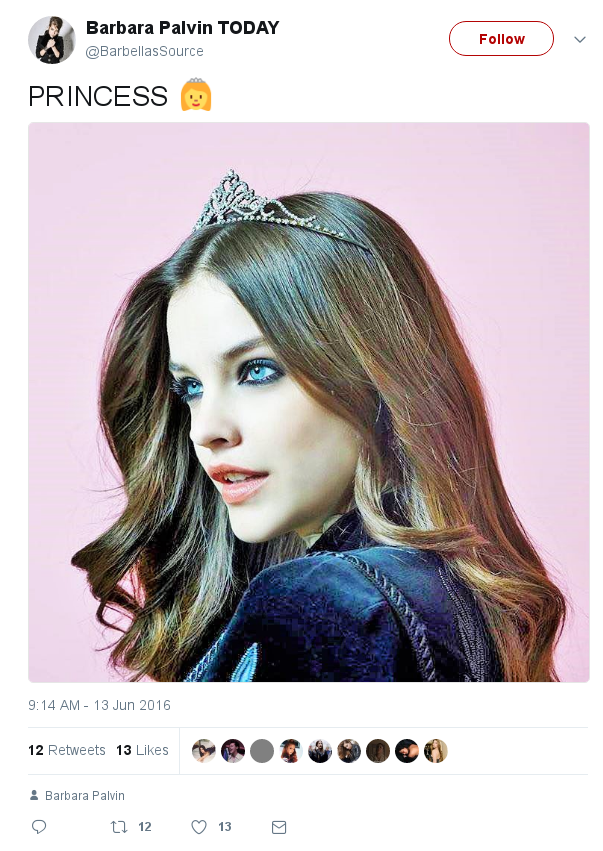}\\
D. & E. & F. \\
\includegraphics[width=0.25\textwidth]{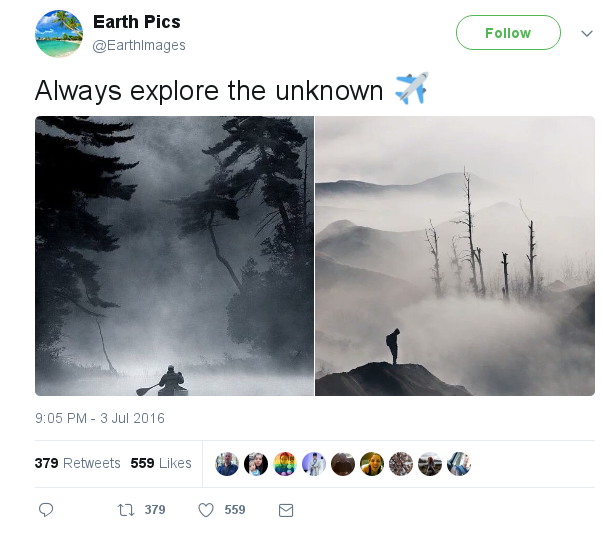} & \includegraphics[width=0.25\textwidth]{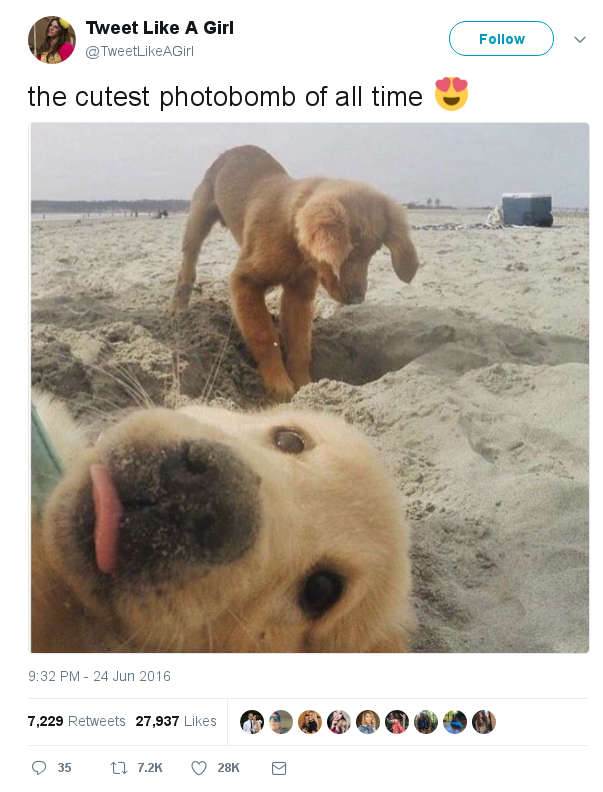} & \includegraphics[width=0.25\textwidth]{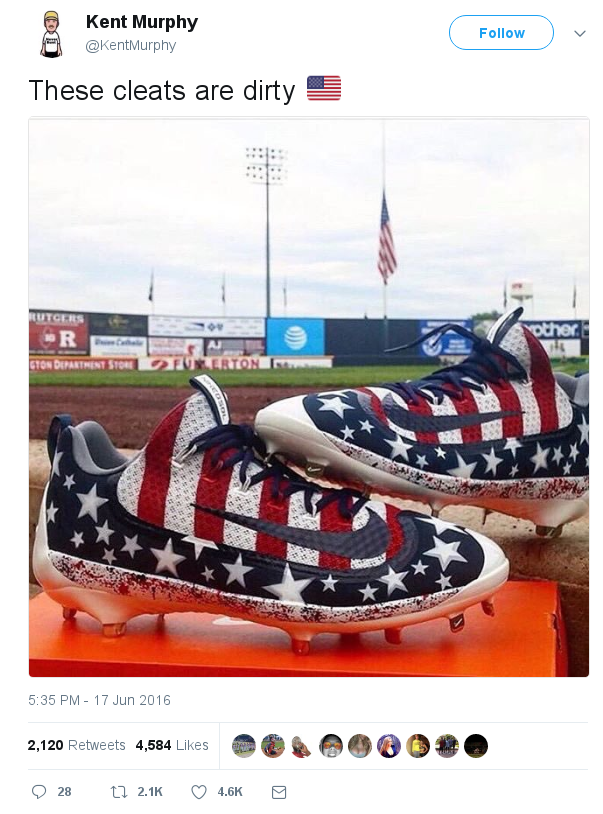}\\
G. & H. & I. \\
\end{tabular}
\caption{Example tweets from the proposed Twemoji dataset. Emoji are removed and used as ground truth annotation. The top row gives examples of text-only tweets, while the bottom rows contain both the text and image modalities. We see the interactions between the three modalities (text, images, and emoji) can vary. For example, F has a strong alignment between all three, while the correlation between the emoji and the tweet is more obvious in the image than the text. Sometimes emoji re-confirm content, as in E, and sometimes they express a sentiment as in D. G gives an example where the emoji modify the content semantics -- the airplane emoji adds a suggestion of travel that is not strictly present in either the text or image modalities. Emoji are intertwined with their related modalities, but are definitely not subsumed by them.}
\label{fig:datasetexamples}
\end{figure*}

\subsubsection{Query-by-Emoji - Can we query with emoji?}

Just as relevant emoji can be suggested for given input modality, they can instead be used as the query modality. Emoji have some unique advantages for retrieval tasks. The limited nature of emoji (1000+ ideograms as opposed to 100,000+ words) allows for a greater level of certainty regarding the possible query space. Furthermore, emoji are not tied to any particular natural language, and most emoji are pan-cultural. This means that emoji can be deployed as a query language in situations where a spoken language might fail. For example, with children who haven't yet learned to read, or perhaps even high intelligence animals such as apes. Further, the square form factor of emoji works naturally with touch screen interfaces. Many of these advantages are shared by any ideogram scheme, but emoji have the additional benefit of exceptional cultural penetration. Because emoji are already adopted and used daily by millions, the cognitive burden to learn what emoji are available to use as queries is significantly decreased. 



In the Query-by-Emoji challenge, we aim to quantify performance on the task of multimedia retrieval given an emoji query. Samples in the test set should be ranked by the model for a given emoji query, and performance will be evaluated based on whether those documents are considered relevant to that emoji or not. 

\subsection{Dataset}

\begin{table}
\caption{Twemoji Dataset and Subset Statistics. Full is the entire collection, Balanced has a class-balanced test set but uses the same training and validation sets, and Images is composed from those tweets with attached images.}
\label{tab:dataset}
\centering
\begin{tabular}{l r r r}
\toprule
				 & Full & Balanced & Images\\\midrule
\# Train Samples &  13M & -- & 917K\\
\# Validation Samples   &   1M & -- &  80K\\
\# Test Samples  &   1M &  10K  &  80K\\
\# Emoji Present & 1242 & 1242 & 1082\\
\bottomrule
\end{tabular}
\end{table}

To facilitate research on these challenges, it is necessary to use a dataset with sufficient examples of the relationship between emoji and other modalities. 
Existing works on emoji have either forgone the use of an annotated emoji dataset or have used datasets comprised of only a small subset of available emoji. 
Both of these settings are artificial and fail to adequately represent the challenge and promise of emoji. 
Instead, we target the full range of potential emoji, including their very long tail, and seek to learn their real-world usage rather than place any prior assumptions on them. 
We construct our dataset, which we call Twemoji, from the popular microblogging platform Twitter, and also identify two valuable subsets of the dataset. The dataset and details of the splits discussed below are publicly available.\footnote{Twemoji Dataset, \href{https://doi.org/10.21942/uva.5822100}{DOI: 10.21942/uva.5822100}}

\label{sec:data}

\begin{figure}
\centering
\includegraphics[width=0.75\columnwidth]{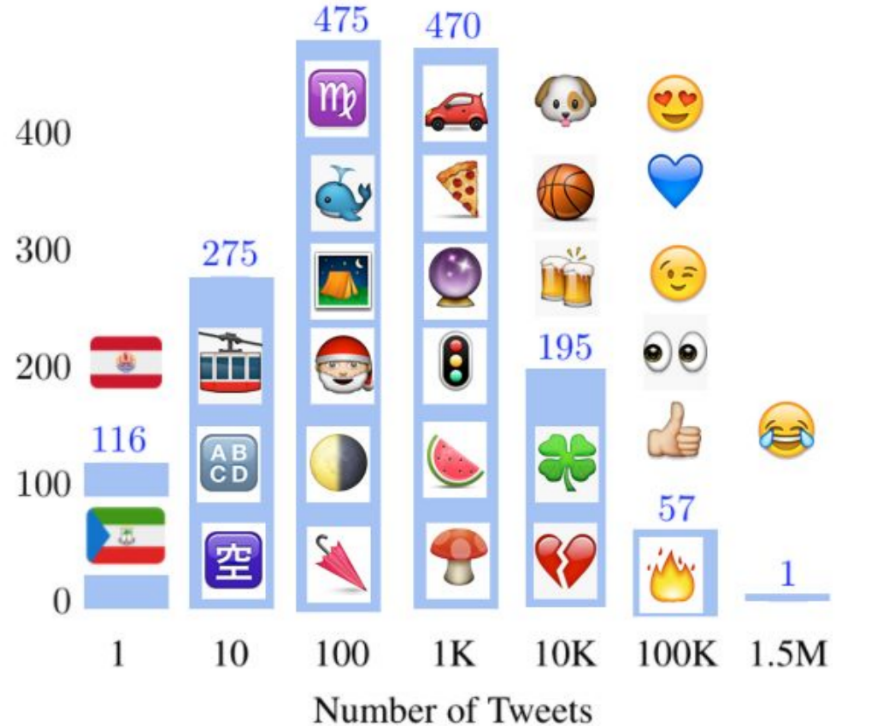}
\caption{Emoji Usage Histogram. The bars show the count of emoji appearing in at least N tweets---\eg, 275 different emoji each appear in 10-100 tweets. In each column, some examples of the emoji in that rarity bracket are displayed.}
\label{fig:emoji_histogram}
\end{figure}

To generate a representative emoji dataset, we collected 25M tweets via the Twitter streaming API during the summer of 2016, filtering these to 15M unique English language tweets that contain at least one emoji. Figure \ref{fig:datasetexamples} gives some examples of tweets in our dataset. Emoji are common on Twitter, appearing in roughly 1\% of the tweets posted during our collection period. However, the usage frequency is heavily skewed (see Figure \ref{fig:emoji_histogram}). \emoji{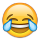} is the most commonly used emoji, and it appears in 1.57M tweets. The top emoji (appearing in 100K+ tweets) are mostly facial expressions, hearts, and a few hand gestures (\emoji{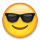},\emoji{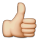},\emoji{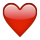}). Most emoji in the dataset have only hundreds (\emoji{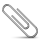}, \emoji{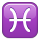}) and thousands (\emoji{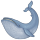}, \emoji{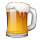}) of examples. Flags and symbols compose the bulk of the rarer emoji.


A fraction of the tweets also contain images, which allow us to present results for the relationship between not only text and emoji but also images and emoji. We therefore present three selections of this dataset: Full, comprised of all tweets in the collection; Balanced, which has a test set constructed with a flattened distribution across emoji; and, Images, which is comprised of those tweets in the collection containing both emoji and images. We present statistics for the three subsets in Table \ref{tab:dataset}, and describe their composition below.

\subsubsection{Twemoji (Full)}

\begin{figure}
\includegraphics[width=\columnwidth]{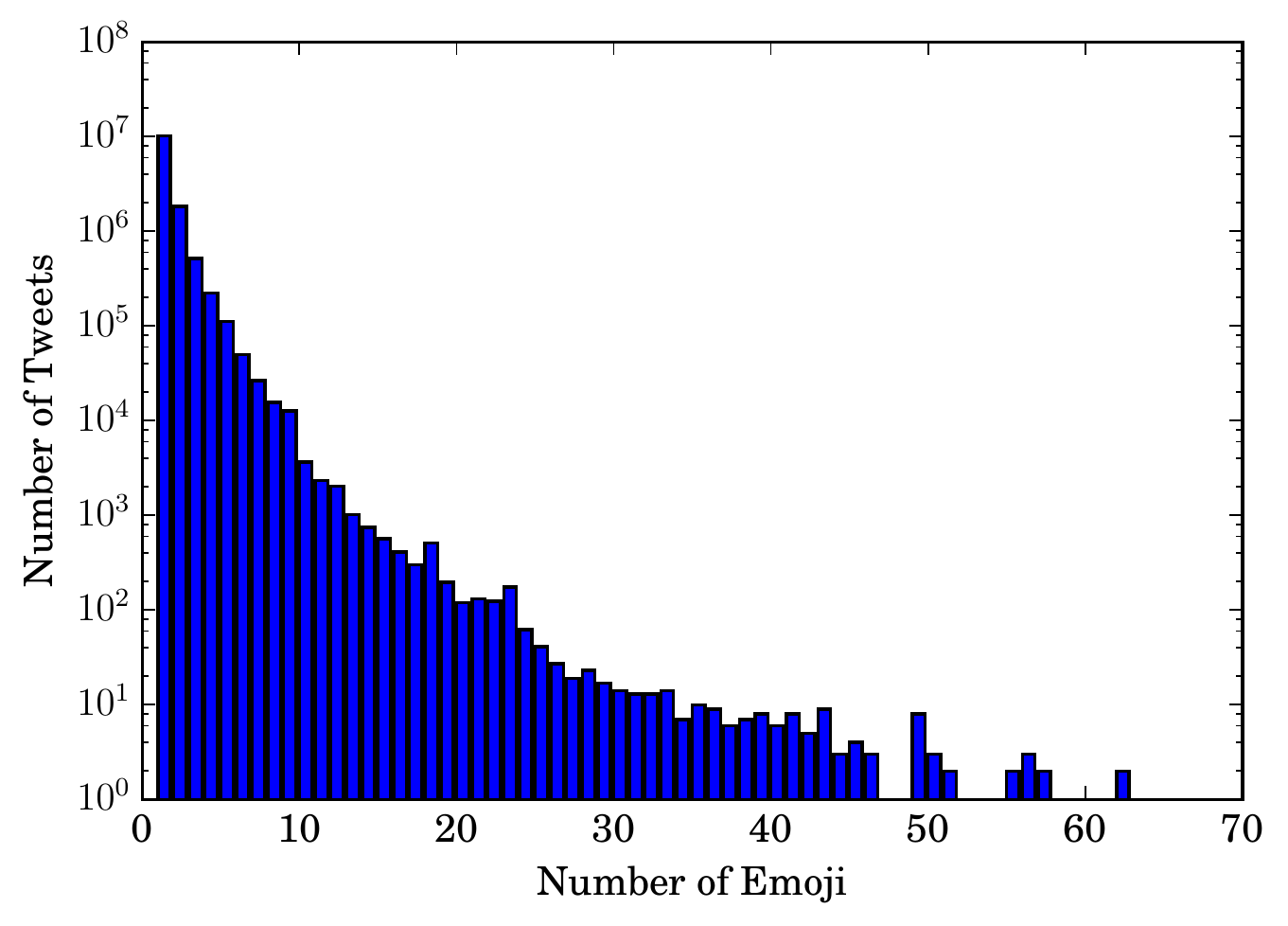}
\caption{Frequency of tweets containing multiple, distinct emoji in the Twemoji-Full training set, plotted on a log scale. We see that a few tweets contain many emoji, but the majority of tweets contain only one or two different emoji.}
\label{fig:multilabel}
\end{figure}

The Twitter data set is split randomly into training, validation, and test sets containing 13M, 1M, and 1M tweets, respectively. Input and annotation pairs are created by removing the emoji from the tweets' text to use as annotation. This approach means that the data set is multi-label, though the predominance of tweets have only one correct annotation. 
Figure \ref{fig:multilabel} shows the number of tweets with a given emoji annotation count. Noting that the $y$-axis is plotted on a log scale, we see that there are almost an order of magnitude more tweets with one emoji than with two emoji, and the numbers continue to drop. A few tweets contain very many emoji. These are perhaps tweets where emoji are being used as a visual language.

The use of emoji as annotation assumes that the majority of emoji provide only supplementary information, and are not operating merely as one-to-one replacements for text tokens (\eg, ``in \emoji{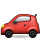} going to \emoji{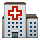} to meet new \emoji{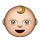}'' is no longer parseable text without the emoji, while for ``awesome day \emoji{cool_shades}" the message remains complete without the emoji). 


\subsubsection{Twemoji Balanced}

We assume that current emoji interfaces may be a contributing factor to the distribution skew of emoji usage. The difficulty in navigating to a desired emoji, compounded with users being unfamiliar with rarer emoji, means that the heavy skew of the distribution could be a self-fulfilling prophecy and an undesired one. Further, it is not clear that the skew of commonly used emoji says anything about their relevance for new tasks like summarization using emoji. We therefore target the case when all emoji are used equally often. Targeting an equal balance ensures that commonly overlooked emoji will still be suggested, and can help eliminate undesired dataset biases. To evaluate this, we test on a more balanced, randomly selected subset of the test set in addition to the full, unbalanced test set. 

The balanced subset is selected such that no single emoji annotation applies to more than 10 examples. To train toward this objective while still leveraging the breadth of the available data, we construct our mini batches so that each emoji has an equal chance of being selected. Namely, the likelihood of selecting a particular sample $x_i$ is
\begin{align}
p(x_i) = \frac{C(y_i)^{-1}}{\sum C(y_i)^{-1}}
\end{align}
where $C(y_i)$ returns the total count of samples with the same emoji annotation $y_i$. While over time this assures that every emoji equally contributes to the model updates, the model will still gain a more nuanced understanding of the more common emoji due to the diversity of their training samples.

\subsubsection{Twemoji Images}

Not all of the images contained in the tweets were still available on the internet, but those that were were downloaded. From these, we constructed a subset of the dataset for which both image and text inputs were available. Due to the prevalence of image-sharing on Twitter and the internet as a whole, a large number of tweets contain the exact same image as other tweets. We use the image-bearing tweets in the full Twemoji test set as our test set. We allow duplicate images between the train and test sets, but only when the emoji annotation of the test set differs from that in the training set. This results in a training set of 900k images, and validation and test sets of 80k images.

\subsection{Evaluation Protocols}


\subsubsection{Emoji Prediction} 

Performance in the Emoji Prediction challenge is reported in both Top-$k$ accuracy and mean samplewise Average Precision (msAP). Top-$k$ accuracy corresponds directly to the scenario in which a system is suggesting some $k$ emoji that the user may wish to include during message composition, and the system should try to ensure that at least some of these emoji are relevant. As our dataset is multi-label, we calculate Top-$k$ accuracy by considering a prediction as correct if \emph{any} predicted class in the top $k$ is annotated as relevant, and a prediction as false if there are none. This means that an emoji ranking for a given input may score a relevant emoji as very unlikely, but still be marked as correct if a different, relevant emoji is correctly predicted in the top $k$. For $N$ samples, where each input $x_i$ has a corresponding binary vector $y_i$ indicating emoji relevancy, the top-$k$ accuracy is calculated with
\begin{align}
\text{ind}_k(x_i, y_i) = 
\begin{cases}
1 & \sum_{j\in p(y_i|x_i)_k} y_i^j > 0\\
0 & \text{otherwise}
\end{cases}
\end{align}
\begin{align}
\text{Top-}k = \frac{\sum_{i=0}^N  \text{ind}_k(x_i, y_i)}{N}
\end{align}
where $p(y_i|x_i)_k$ yields the indices of the $k$ highest scoring class predictions, and $y_i^j$ corresponds to the value of the $j$th element of $y_i$.

To offer a more complete picture, we also report the mean samplewise Average Precision. This measures the performance of the algorithm across the entire ranking of emoji for a given input. It evaluates how accurately ranked the emoji are for a given image and/or text input.
%
\begin{align}
\text{msAP} = \frac{1}{N}\sum_i^N \frac{\sum_j^C Prec(j)\times y_i^j}{\sum y_i}
\end{align}
where $Prec(j)$ gives the precision of the prediction at rank $j$, and $y_i^j$ gives the value of $y_i$ at the index $j$.


\subsubsection{Emoji Anticipation} 
%
Emoji Anticipation differs from Emoji Prediction in its absence of training data, but the test set and goal of the challenge is shared with Emoji Prediction. For this reason, results are again reported in both Top-$k$ accuracy and msAP.

\subsubsection{Query-by-Emoji}
Query-by-Emoji turns the problem on its head: given a query emoji, the goal is retrieve a ranked list of documents considered relevant due to their text or image content. As this corresponds to a more classical retrieval problem, we report results in mean Average Precision (mAP) across all single emoji queries
\begin{align}
\text{mAP} = \frac{1}{C}\sum_i^C \frac{\sum_j^N Prec(j)\times y_j^i}{\sum_j y_j^i}
\end{align}
where $C$ is the number of single emoji queries, $N$ is the number of samples, and $y_j^i$ corresponds to the relevancy of query $i$ to the $j$th ranked sample.

\section{Emoji Prediction}
\label{sec:supervisedprediction}

\subsection{Baselines}
\subsubsection{Text-to-Emoji}
\label{sec:text2emoji}

Our baseline text model consists of a bi-directional LSTM, which processes the text both in standard order and reverse order, on top of a word embedding layer \cite{mikolov2013efficient}. LSTMs use their memory to help emphasize relevant information~\cite{hochreiter1997long}, but there is still a degradation of information propagation. The bi-directional nature of the LSTM helps to combat this effect and ensure that information from the beginning of the sentence isn't lost in the representation.

Words are placed in a vector embedding space, passed through our bi-directional LSTM layers, and the resultant representations are combined and fed to a softmax layer that attempts to predict relevant emoji. Text from the Twemoji dataset is tokenized and used to train the model. The validation set is used to determine after how many epochs to stop training (to avoid overfitting).

%


\subsubsection{Image-to-Emoji}

Similar to the approach for text-based prediction, we can also train a model for image-to-emoji prediction using our data. 
We use a CNN to represent images accompanying tweets. It is a GoogLeNet architecture trained to predict 13k ImageNet classes \cite{googlenet, mettes2016imagenet}. 
We use the representation yielded at the penultimate layer for our image input. 
We train a single soft-max layer on top of this representation with emoji prediction as the objective, with the weights prior to this softmax frozen. 
An end-to-end convolutional model could also be trained with sufficient training data, but it would be difficult to amass the requisite number of training samples, particularly for the longtail of the emoji usage distribution. 

\subsubsection{Fusion}

For the combination of both text and image modalities, a late fusion approach is used. As both the text-based neural network and the image-based convolutional network output emoji confidence scores in a softmax layer, their format is directly comparable. Given confidence scores $p_{txt}(y|x_{txt})$ predicting the likelihood of a given emoji $y$ for some text $x_{txt}$ and the corresponding scores $p_{img}(y|x_{img})$ for some image $x_{img}$, we give a combined prediction:
\begin{align}
p(y|x_{txt},x_{vis}) = \alpha p_{txt}(y|x_{txt}) + (1-\alpha) p_{img}(y|x_{img})
\end{align}

where $\alpha$ is a modality weighting parameter in the range $[0,1]$ which is determined through validation.

\subsection{Results}

\begin{table}
\caption{Results for text-based emoji prediction. Though not directly comparable, we observe stronger performance on the balanced test set. This is expected behaviour as we targeted the balanced likelihood during training.}
\label{tab:suptxtresults}\centering
\begin{tabular}{l r r r r r}
\toprule
\bfseries Dataset & \bfseries Top-1 & \bfseries Top-5 \bfseries & \bfseries Top-10 & \bfseries Top-100 & \bfseries msAP \\\midrule
Twemoji (Full)	& 13.0 				& 30.0			& 41.0		&	84.0 & 19.4	\\
Twemoji-Balanced 	& 35.1 &  48.3& 54.7 			& 	87.7		& 35.1 \\
\bottomrule
\end{tabular}
\end{table}

\begin{figure}
\includegraphics[width=\columnwidth]{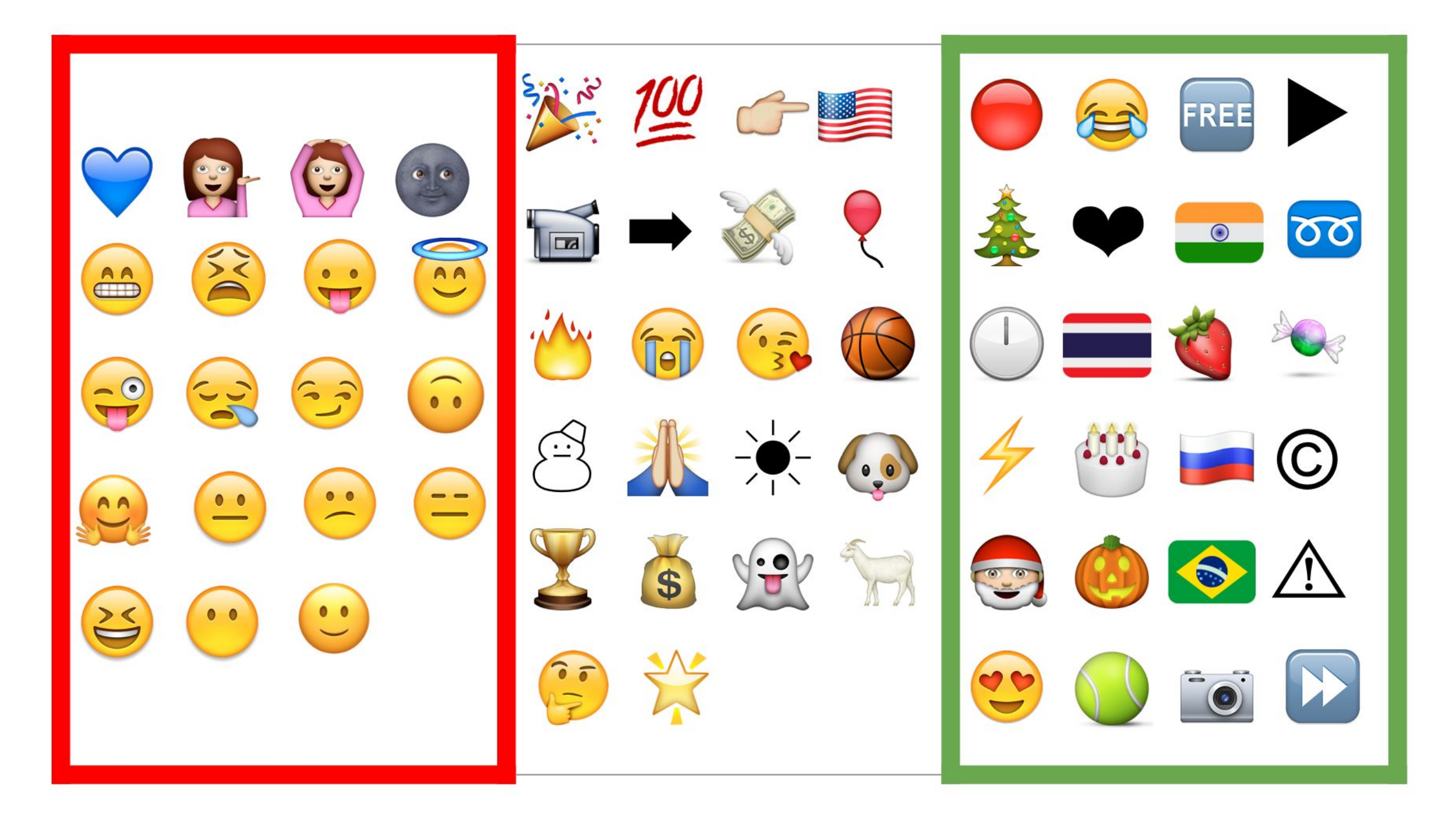}
\caption{Examples of the hardest emoji to predict (red), the easiest (green), and those in between. Ambiguous faces are difficult to predict, while emoji tied concretely to an event, object, or place tend to be the easiest.}
\label{fig:easymediumhard}
\end{figure}

\subsubsection{Text-to-Emoji}

The results for prediction on the Twemoji test sets are shown in Table \ref{tab:suptxtresults}. 
Figure \ref{fig:easymediumhard} gives examples of those emoji the baseline models find difficult or easy to predict. We see that some of the most difficult emoji to predict include ambiguous face emoji where no clear emotion is displayed. Among the easiest emoji to predict are flag emoji and emoji tied closely to particular events, such as Christmas or birthdays. We also see less obvious emoji such as \emojipng{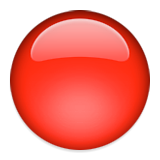} included. This is likely due to the resemblance of \emojipng{redcircle} to a recording symbol on a video camera, as it is often used in conjuction with tweets containing links to video. It is likely this co-occurrence that makes it a particularly easy emoji to predict. Such usage underscores the necessity of using real world emoji usage where possible, as the unicode name for \emojipng{redcircle} is merely `Large Red Circle' which gives little to relate it to video.

\begin{table*}
\caption{Results of the CNN-based image-input model and the bi-directional LSTM text-input model on Twemoji-Images, as well as the fusion of the two.}
\label{tab:supimages}
\centering
\begin{tabular}{l l r r r r r }
\toprule
& \bfseries Model & \bfseries Top-1 & \bfseries Top-5 & \bfseries Top-10 & \bfseries Top-100 & \bfseries msAP \\
\midrule
\multirow{2}{*}{Single Modality} & Image only & 14.7 & 33.0 & 44.0 & 86.4 & 17.0  \\ 
& LSTM (Text Input) & 17.7 & 33.5 & 43.4 & 81.3 & 22.3 \\
\midrule
Fusion & Image + LSTM ($\alpha = 0.6$) & \bfseries 20.6  &\bfseries 40.3 &\bfseries 51.5 &\bfseries 89.3 &\bfseries 27.0 \\
\bottomrule
\end{tabular}
\end{table*}

\begin{table*}
\centering
\caption{Examples of text-to-emoji and image-to-emoji prediction results on the Twemoji-Images test set. We observe that sometimes images or text capture important predictive content that isn't present in the other modality, and sometimes both modalities fail to yield the expected emoji. In general, few of the suggested emoji seem unreasonable from a subjective standpoint, which suggests that perfection on the evaluation metrics is not necessary for the models to be usable.}
\label{tab:qualitativeexamples}
\centering
\begin{tabularx}{.9\textwidth}[t]{c c p{2in} c c c}
\toprule
& \bfseries Image & \bfseries Text & \bfseries Image-only & \bfseries Text-only & \bfseries True Emoji\\\midrule
A & \raisebox{-.5\height}{\includegraphics[height=0.5in]{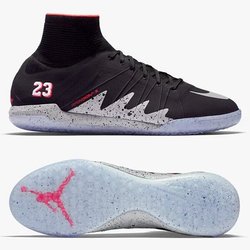}} & rt U : nah this neymar x jordan collab is pure heat & \emojipng{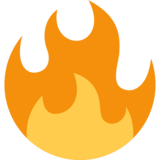}~\emojipng{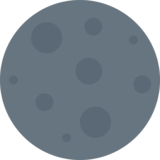}~\emojipng{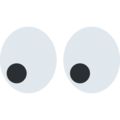}~\emojipng{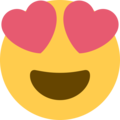}~\emojipng{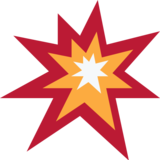} &\emojipng{fire}~\emojipng{eyes}~\emojipng{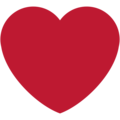}~\emojipng{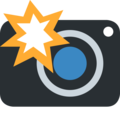}~\emojipng{smiling-face-with-heart-shaped-eyes} & \emojipng{fire}\\
 & & & & \\
B & \raisebox{-.5\height}{\includegraphics[height=0.5in]{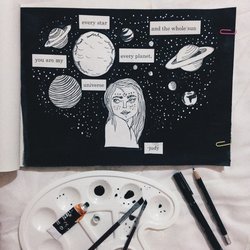}} & rt U : one of the short poetry i have done , \#watercolor \#art & \emojipng{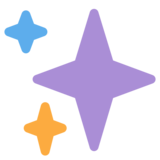}~\emojipng{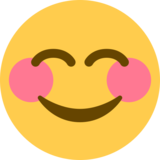}~\emojipng{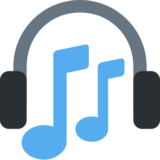}~\emojipng{milky-way}~\emojipng{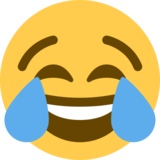} & \emojipng{face-with-tears-of-joy}~\emojipng{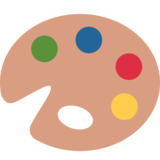}~\emojipng{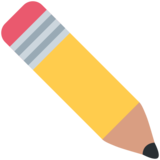}~\emojipng{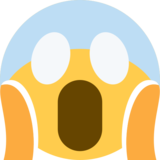}~\emojipng{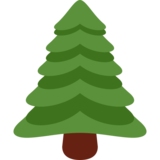} & \emojipng{milky-way}\\
 & & & & \\
C & \raisebox{-.5\height}{\includegraphics[height=0.5in]{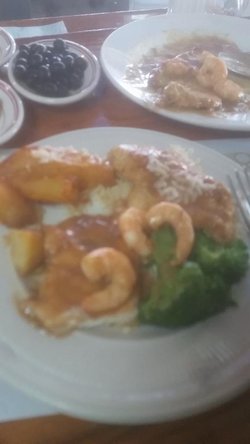}} & thank you & \emojipng{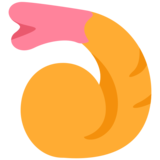}~\emojipng{face-savouring-delicious-food}~\emojipng{smiling-face-with-smiling-eyes}~\emojipng{smiling-face-with-heart-shaped-eyes}~\emojipng{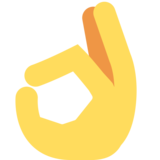} & \emojipng{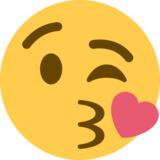}~\emojipng{smiling-face-with-heart-shaped-eyes}~\emojipng{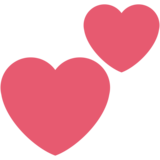}~\emojipng{smiling-face-with-smiling-eyes}~\emojipng{heavy-black-heart} & \emojipng{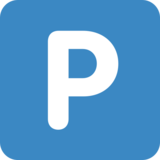}~\emojipng{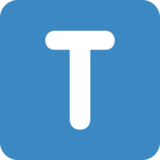}\\

 & & & & \\
D & \raisebox{-.5\height}{\includegraphics[height=0.5in]{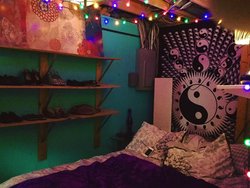}} & turned my ghetto concrete workshop room into my own cool little space& \emojipng{face-with-tears-of-joy}~\emojipng{smiling-face-with-heart-shaped-eyes}~\emojipng{smiling-face-with-smiling-eyes}~\emojipng{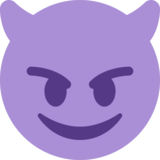}~\emojipng{eyes} & \emojipng{fire}~\emojipng{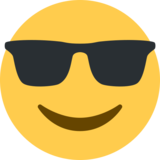}~\emojipng{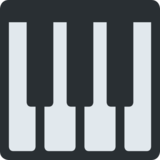}~\emojipng{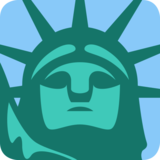}~\emojipng{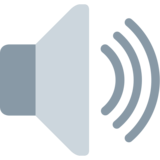}~\emojipng{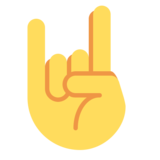} & \emojipng{smiling-face-with-sunglasses}\\
 & & & & \\
E & \raisebox{-.5\height}{\includegraphics[height=0.5in]{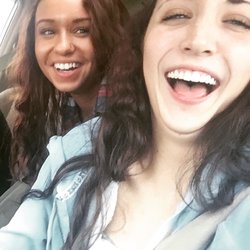}} & no one will ever understand what it's like to have a best friend like this so lucky i am U & \emojipng{heavy-black-heart}~\emojipng{two-hearts}~\emojipng{face-throwing-a-kiss}~\emojipng{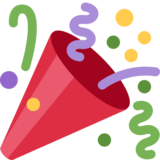}~\emojipng{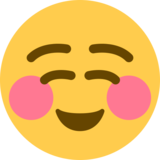} & \emojipng{heavy-black-heart}~\emojipng{smiling-face-with-heart-shaped-eyes}~\emojipng{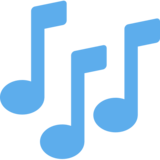}~\emojipng{two-hearts}~\emojipng{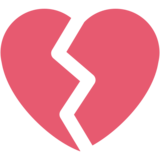} & \emojipng{heavy-black-heart}~\emojipng{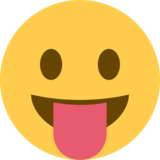}~\emojipng{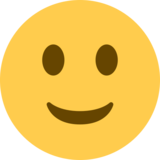}\\
 & & & & \\
F & \raisebox{-.5\height}{\includegraphics[height=0.5in]{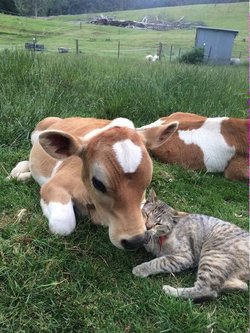}} & not food & \emojipng{smiling-face-with-heart-shaped-eyes}~\emojipng{heavy-black-heart}~\emojipng{smiling-face-with-smiling-eyes}~\emojipng{white-smiling-face}~\emojipng{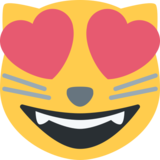} & \emojipng{fried-shrimp}~\emojipng{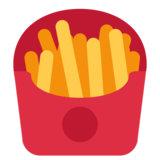}~\emojipng{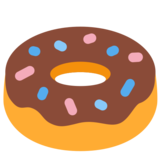}~\emojipng{hamburger}~\emojipng{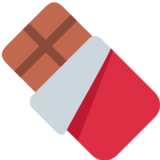} & \emojipng{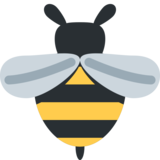}\\
 & & & & \\
G & \raisebox{-.5\height}{\includegraphics[height=0.5in]{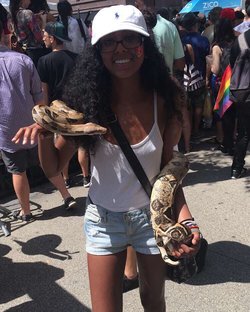}} & im that weird girl that likes to hold snakes & \emojipng{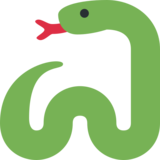}~\emojipng{smiling-face-with-heart-shaped-eyes}~\emojipng{heavy-black-heart}~\emojipng{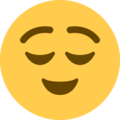}~\emojipng{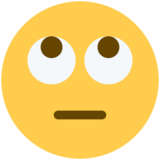} & \emojipng{snake}~\emojipng{put-litter-in-its-place-symbol}~\emojipng{face-with-tears-of-joy}~\emojipng{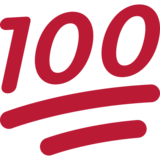}~\emojipng{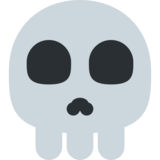} & \emojipng{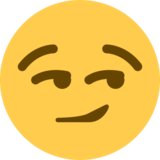}\\

\bottomrule
\end{tabularx}
\end{table*}

It is worth noting that the numbers here reflect accuracy on predicting the emoji that \emph{were} used, which are not necessarily all the emoji which could have been used. It is likely that some emoji were predicted which could be argued as relevant but which happened to not be the particular emoji the Twitter user selected. While the results should be considered indicative, the annotations used cannot be considered absolute due to the subjectivity of emoji.

We note that the model performs much stronger on the balanced dataset. This is expected, as we targeted a balanced distribution during training, due to the assumption that some amount of the data bias was due to intrinsic bias in input interfaces. While we target a balanced distribution, the model can also be trained without balanced sampling to learn the skewed distribution. The model, when trained without balanced sampling achieves top-1 accuracy of 21.4\% and 19.9\% on the raw and balanced test sets, respectively. From a practical standpoint, this is a far less interesting result due to the heavy skew in data. While this greatly improves the performance on the raw test set, the performance on the balanced subset diminishes significantly. We restrict all further discussion to only models that have been trained with a balanced sampling regime.

\subsubsection{Image-to-Emoji}

As described previously, we train a model to predict emoji based on CNN representations of images. In the top section of Table \ref{tab:supimages}, we present the results of the image-trained model on the available image-bearing test set. We also present results for testing the text-trained model on this subset. We see that the image modality is competitive to the text modality for the prediction of emoji. This suggests that the emoji may often be as related to the images as they are to the text content. Overall, the performance of the models are broadly similar to those on the full Twemoji dataset, which is encouraging. It suggests that the relationship between the input data and the annotation is not too dissimilar to the whole set in this subset.

Table \ref{tab:qualitativeexamples} gives some qualitative examples of results for emoji prediction on image and text inputs, along with the ground truth emoji annotation. Example C captures the food aspect of the image which is missed in the text modality, but neither are able to predict the true emoji. This is an example where the information contained in the emoji modality is mostly orthogonal to that in the text or image. We see in example F that the text-based prediction is led astray by the mention of food while the image-based method focuses on the emotional reaction expected from cuddling animals. The correct emoji, \emojipng{honeybee}, appears in the top 100 results for the image-based baseline, while it is in the 400s for the text modality. Some examples are easily handled by both the text and image modalities, such as A -- this may be due to a strong association between the \emojipng{fire} emoji and sneaker enthusiasts. Example B is an interesting one, because both the image and the text contained the context of artwork, but the image was able to retrieve the artwork's content and associate it with the correct emoji \emojipng{milky-way} while that content was not available in the text. 


\subsubsection{Fusion}
\begin{figure}
\centering
\includegraphics[width=0.9\columnwidth]{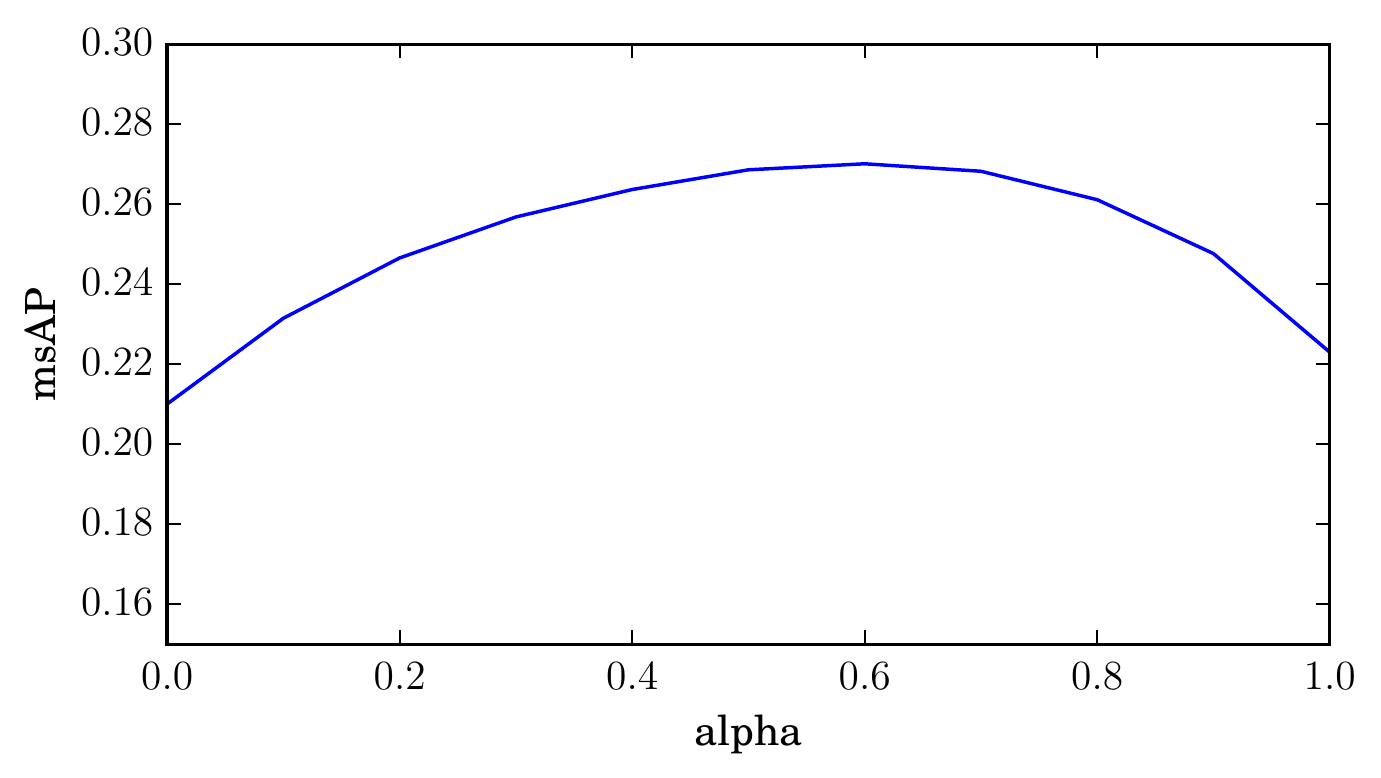}
\caption{Effect of modality-weighting parameter $\alpha$ on the prediction of Twemoji-Images, measured in mean samplewise Average Precision. $\alpha=1.0$ corresponds to using only the text predictions, while a value of $0.0$ corresponds to using only image predictions. Peak performance occurs near $\alpha=0.6$. The overall improvement through combining both modalities tells us that the modality streams have complementary information for the prediction of emoji. }
\label{fig:supervisedfusion}
\end{figure}

In the bottom of Table \ref{tab:supimages}, we provide scores for a fusion of both the image and text modalities. We see a significant improvement across most metrics through the fusion of both modalities, which tells us that they have complementary information. Though this could be an artifact of the representations used in either modality, it is reasonable to assume that the semantics of the emoji are not strictly tied to either modality, which is evidence that emoji should be considered as a modality in their own right. In Figure \ref{fig:supervisedfusion}, we show the per-sample mAP (ranking emoji given an image+text input) performance as a function of the fusion weighting parameter $\alpha$. We see that the curve hits its peak near the center, with a skew toward the text input. This suggests a slightly stronger correlation between the emoji modality and text than between emoji and images. 

\begin{figure}
\centering
\includegraphics[width=.9\columnwidth]{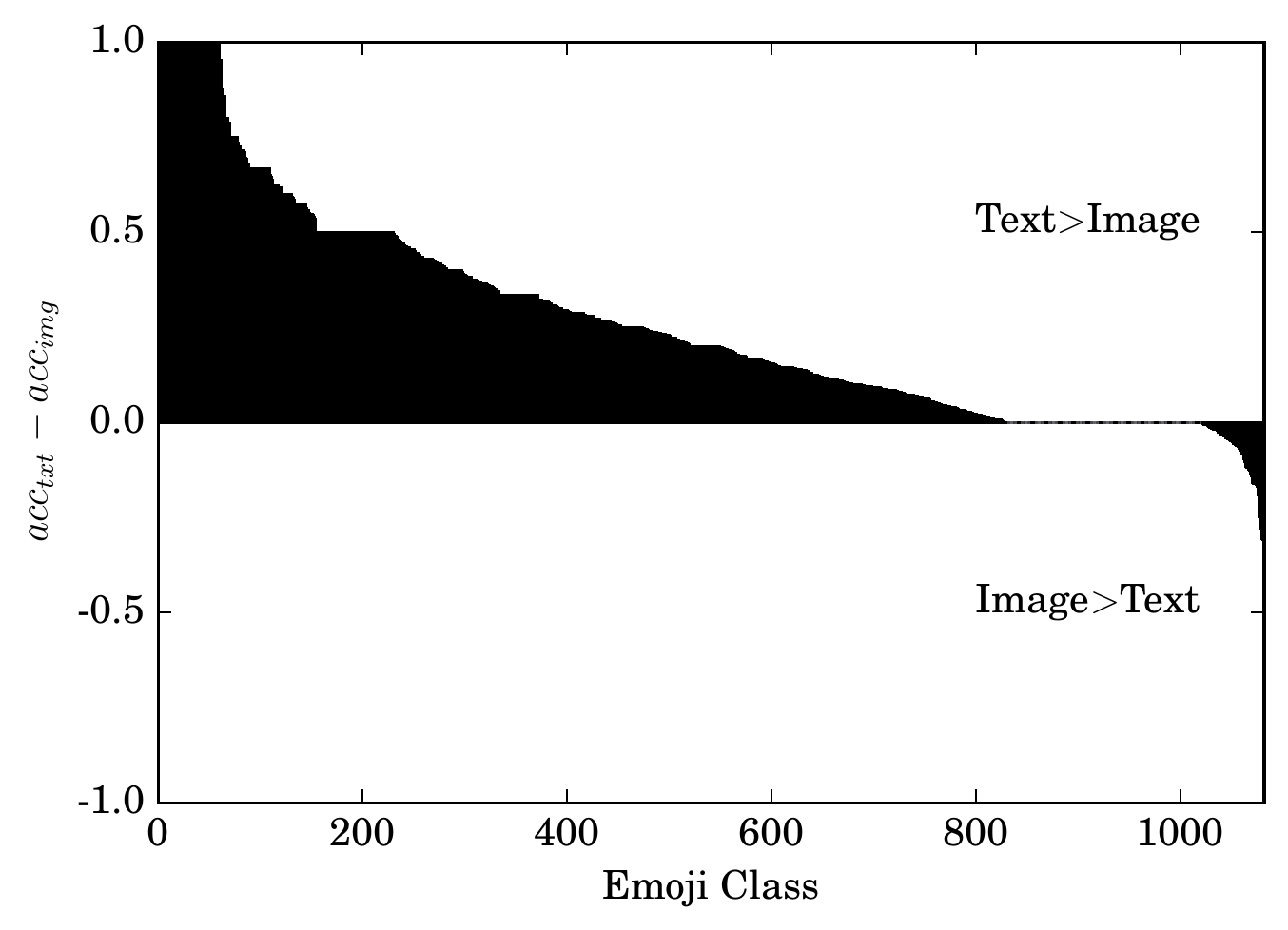}
\caption{Per-class performance difference between text and image modalities. This graph shows the difference in Top-$5$ accuracy between using solely the text input modality to predict emoji and using solely the image input modality. For roughly 80\% of the emoji, text outperforms images for our dataset and baselines.}
\label{fig:textvsimages}
\end{figure}

In Figure \ref{fig:textvsimages}, we report the per-class difference in the msAP metric. This difference is calculated by subtracting the image-based performance from the text-based performance. A value of 0.0 would therefore mean that both methods performed identically well (or poorly), a positive value indicates that the text-based model performed better, and a negative indicates that the image-based model performed better. A strong bias toward the text-based approach is observed across almost all emoji. It is impossible to say whether this reflects the strength of cross-modal affinities, but it does tell us that the model we use for relating text to emoji is stronger than that for images. 

\section{Emoji Anticipation}
\label{sec:zeroshotemoji}

\subsection{Baselines}
\subsubsection{Text- and/or Image-to-Emoji}

Word embeddings have been used for the task of zero-shot image classification as a means to transfer knowledge from one class to another \cite{conse}. 
To place an emoji within this embedding space without the need for training examples, a short textual description of the emoji can be used as its representation. 

We utilize a word2vec representation~\cite{word2vec} that is pre-trained on a corpus of millions of lines of text accompanying Flickr photos~\cite{yfcc100m}. 
Input modalities are then embedded in this shared space, where relationships between items are evaluated by their similarity in the space. 
Text terms are placed directly in the space through vocabulary look-up, as the embedding is originally trained on text. 
In the case of images, the names of the highest scoring visual concepts are used, weighted by their confidence scores. 
We use 13k visual concept scores that come from the same GoogLeNet-style CNN used to extract high level features in the supervised setting. 

To place the emoji modality within this mutual vector space, we use text terms extracted from the unicode-specified emoji title and descriptions. Emoji are unicode characters, and the details of their illustration are left to the implementation of the platform which incorporates them. However, when new emoji are accepted into the unicode specification, they are presented with a title and description. We take the averaged word2vec vector representation of the words in this specification as a vector representative of that emoji within our space.

For emoji prediction using a fusion of text and image inputs, we use a simple weighted late fusion approach in the manner described in the previous section. Because we don't have any validation (or training) data in the unfamiliar emoji setting, the weighting parameter $\alpha$ cannot be experimentally determined. Instead, we assign $\alpha=0.5$, giving both text and visual modalities equal priority in our model. 

\begin{table}
\centering
\caption{Emoji Anticipation results, reported on Twemoji-Images. Emoji are predicted without any direct supervision data, analogous to what must be done when new emoji are released. We see improvement across all metrics when a fusion of the input modalities is used.}
\label{tab:zeroshotresults}
\begin{tabular}{ l r r r r r }
\toprule
\bfseries Model & \bfseries Top-1 & \bfseries Top-5 & \bfseries Top-10 & \bfseries Top-100 & \bfseries msAP\\\midrule
Random & 0.0 & 0.4 & 0.9 & 8.1 & 0.5 \\
Zero-shot Text & 1.1  & 2.5 & 3.9 & 20.9  & 1.9\\
Zero-shot Images & 1.3 & 3.0 & 4.3 & 21.4  & 2.1\\
Fusion ($\alpha=0.5$)& \bfseries 1.5 &\bfseries 3.8 &\bfseries 5.7 &\bfseries 23.8&\bfseries  2.5\\
\bottomrule
\end{tabular}
\end{table}


\subsection{Results}

In Table \ref{tab:zeroshotresults} we give results for emoji prediction on the Twemoji-Images dataset using only the text modality, only the image modality, and the fusion of the two (using $\alpha=0.5$). We observe that, as would be expected, the overall scores are much lower than the supervised approaches in the previous section. Though the results are small, they are significantly above random. The top-$1$ accuracy of random guesses on the Twemoji-Images test set is on the order of  0.08\% compared with 1.5\% for the fusion of the zero-shot results. 

A surprising result is that the Image modality actually outperforms the text modality in most of the metrics. Because the semantic space is learned on textual data, one might expect the text modality to be the most reliably embedded modality within the shared space, but that does not seem to be the case. Perhaps this is a result of many distracting terms in the textual data, which supervised approaches learn to filter out. Meanwhile, the limited vocabulary of the CNN concepts are likely to be a strong signal. Nonetheless, the fusion of the two modalities improves performance across all metrics.

The names of emoji may be reasonable, but might not capture unexpected uses. For example, \emph{fireworks} \includegraphics[height=\mytextsize]{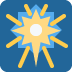} could be used for `north star' or `sun' based solely on its particular illustration here -- usages that would be unlikely to captured based on the title alone. Similarly, \emph{ghost} \includegraphics[height=\mytextsize]{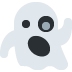} has an especially friendly illustration, with the spectre appearing to wave hello. Such usage based on the visual appearance can easily diverge relative to the drier, more descriptive title.

The performance of this baseline approach can likely be improved by focusing on improving the quality of the mapping of the three modalities to the mutual space. The embedding of emoji, for example, could likely be improved by manually specifying additional relevant text terms. The terms contained in the unicode specification focus on being descriptive about the emoji, focusing on \emph{what} it is, rather than how it might be used. Though difficult to experimentally evaluate in an objective manner, adding some extra terms based on postulated usage to the emoji representation could be one way to boost performance without significant extra effort. For example, \emojipng{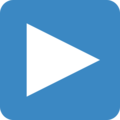} has the title ``black right-pointing triangle", which is a description of what the emoji is but says little about how it might be used. Adding potentially related terms such as \emph{next} or \emph{play} or \emph{therefore} might capture probable usage semantics that are absent in a pure description of the emoji itself. Indeed, due to the particular illustration of this emoji, the term \emph{black} in the description is actually misleading as there is nothing black about the right-pointing triangle in this rendering.

\begin{table}
\caption{Query-by-Emoji results for both supervised and zero-shot baselines. Results are reported in percentage mAP. In the supervised setting, we find the images to slightly outperform the text, but in the zero-shot setting the performance is reversed.}
\label{tab:querybyemoji}
\centering
\begin{tabular}{l r r r}
\toprule
\bfseries Method	& \bfseries Twemoji 	& \bfseries Twemoji 	& \bfseries Twemoji\\
& \bfseries (Full) & \bfseries (Balanced) & \bfseries (Images)\\\midrule
Random 			 	&	0.1					  	&	0.3							& 0.2\\\midrule
LSTM (Text)		 	& 19.3 						& 35.5 							& 20.2\\
CNN (Image) 	 	& -- 						& -- 							& \bfseries22.0 \\
Fusion				& -- 						& -- 							& 21.2\\\midrule
Zero-shot Text 		& 0.5						& 2.0						& 1.5\\
Zero-shot Images 	& -- 						& --							& 0.8\\
Zero-shot Fusion & -- 						& --							& \bfseries1.3\\\bottomrule
\end{tabular}
\end{table}


\begin{table*}
\caption{Top ranked documents for three emoji queries. We see a correspondence between the baseline's prediction of certain emoji and current events, with relationships between \emph{Finding Dory} and the tropical fish emoji, as well as sad current events and the pensive face emoji. Non-relevant results, like those for eyeglasses, may appear subjectively to be relevant but there is clearly a nuance in the usage of the eyeglass emoji that is being overlooked. }
\label{fig:qualqueryresults}
\begin{tabularx}{\textwidth}{c c p{1in} c p{1in} c p{1in}}
\toprule
Query: & \emojipng{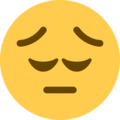} & & \emojipng{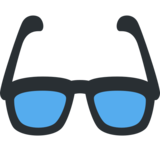} & & \emojipng{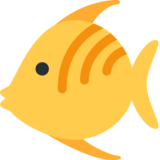}\\\midrule
1 & \raisebox{-.5\height}{\includegraphics[height=0.7in,width=0.7in,keepaspectratio]{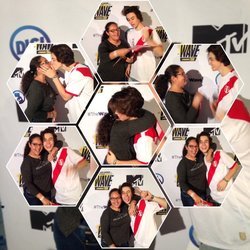}} & you can't imagine how much i miss you \#facetimemenash
& \raisebox{-.5\height}{\includegraphics[height=0.7in,width=0.7in,keepaspectratio]{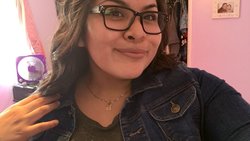}}&rt U : glasses ... no glasses ... glasses
& \cellcolor{green!25}\raisebox{-.5\height}{\includegraphics[height=0.7in,width=0.7in,keepaspectratio]{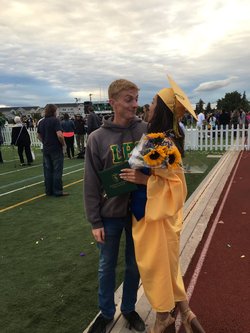}}&\cellcolor{green!25} graduation part N : my favorite fish in the sea
\\
2 & \cellcolor{green!25}\raisebox{-.5\height}{\includegraphics[height=0.7in,width=0.7in,keepaspectratio]{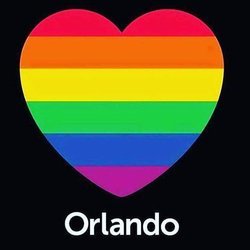}} & \cellcolor{green!25}rt U : so sad \#orlando \#rip
& \raisebox{-.5\height}{\includegraphics[height=0.7in,width=0.7in,keepaspectratio]{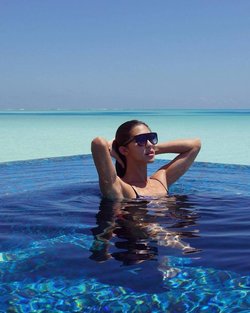}}&rt U : the bigger the better when it comes to eyewear ! by U . london
& \cellcolor{green!25}\raisebox{-.5\height}{\includegraphics[height=0.7in,width=0.7in,keepaspectratio]{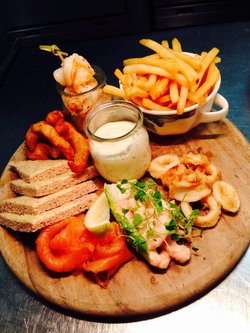}}& \cellcolor{green!25} rt U : it's a fishy kinda day ... fish platter and salmon \& smoked UNKNOWN fish cakes
\\
3 & \raisebox{-.5\height}{\includegraphics[height=0.7in,width=0.7in,keepaspectratio]{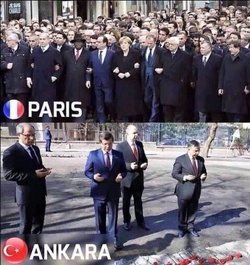}}& rt U : this is so sad \#prayforturkey
& \raisebox{-.5\height}{\includegraphics[height=0.7in,width=0.7in,keepaspectratio]{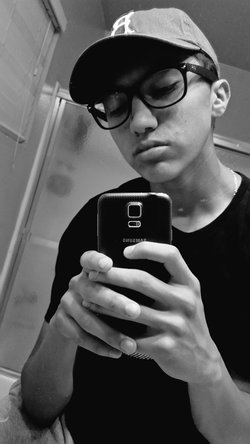}}&rt U : glasses or no glasses
& \raisebox{-.5\height}{\includegraphics[height=0.7in,width=0.7in,keepaspectratio]{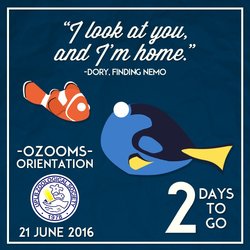}}& N days to go ! just keep swimming swimming swimming UNKNOWN
\\
4 & \raisebox{-.5\height}{\includegraphics[height=0.7in,width=0.7in,keepaspectratio]{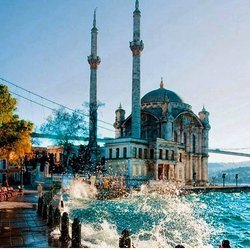}}& my heart goes out to the families and friends who lost their loved ones terrible and sad news ! \#istanbul
& \raisebox{-.5\height}{\includegraphics[height=0.7in,width=0.7in,keepaspectratio]{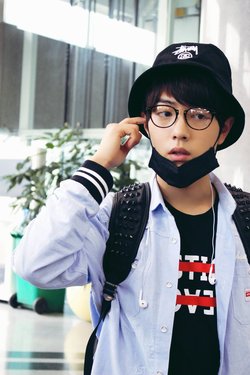}}&glasses
& \cellcolor{green!25}\raisebox{-.5\height}{\includegraphics[height=0.7in,width=0.7in,keepaspectratio]{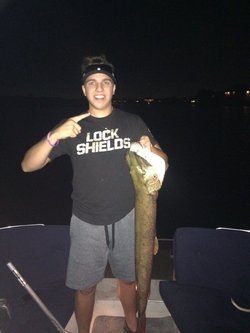}}&\cellcolor{green!25} rt U : i found dory
\\
\bottomrule
\end{tabularx}
\end{table*}

\section{Query-by-Emoji}
\label{sec:querybyemoji}
\subsection{Baselines}
The baselines in previous sections give normalized scores across possible emoji given the input modalities. By calculating these normalized scores for all documents, we are able to rank the documents in order of predicted relevancy to a given emoji query. In this way, we can then perform retrieval per-emoji across these documents. All results in this section are therefore produced by applying the baseline models described in the previous sections to all documents within the test database, and performing retrieval based on per-emoji class scores.

\subsection{Results}

Table \ref{tab:querybyemoji} gives results for the Query-by-Emoji task. Surprisingly, we see that retrieving tweets using only the supervised image understanding slightly outperforms both text-only and the fusion of the two. This result is markedly different from the emoji prediction task where text outperformed images. This could possibly be the result of a very strong correlation within high probability image-emoji pairs. 

In Table \ref{fig:qualqueryresults}, some qualitative query-by-emoji results are shown. We observe strong signals for correlations with current events that occurred during the data collection period of the dataset. Tragic events occurred during this period in both Orlando and Turkey, and the model picked up a strong relationship between the ``pensive face" \emojipng{pensive-face} and these topics. Similarly, the movie \emph{Finding Dory} was released during this time, and we see it present in the high-ranked predictions for the tropical fish. The exploitation and mapping of these emoji-event relationships presents interesting avenues for future research.

For the eyeglasses emoji, the top-ranked results from our baseline model did not contain the eyeglasses emoji. The top four results all contain glasses in the image and a mention of `glasses' or `eyewear' in the text, but the authors opted for alternative emoji during composition. While these results undoubtedly have a level of subjective relevance, the authors clearly felt that other emoji were called for. Perhaps the eyeglass emoji is considered too redundant when the content is already contained in both the text and images. Learning to identify and exploit these subtle distinctions is an open problem for future, improved models.

\section{Conclusion}
\label{sec:conclusion}
In this paper, we have approached emoji as a modality distinct from text and images. There is sufficient motivation for doing so, and considerable future opportunities for research and applications with the emoji modality. We have proposed a large scale dataset of real-world emoji usage, containing the semantic relationships between emoji and text as well as emoji and images. We have defined three challenge tasks with evaluation on this dataset, and provided baseline results for all three. We have looked at the problem of predicting emoji from text and/or images, both with the use of ample training data and in the absence of any. We have also looked at the problem of using emoji as queries for cross-modal retrieval. 
Emoji are everywhere, and are becoming only more pervasive. They already possess a distinct semantic space that can be utilized as a strong information signal as well as a novel means of interaction with data, through both query-by-emoji as well as emoji summarization of content. Furthermore, their semantic richness will only increase as new emoji continue to be introduced. It is our hope that this work and the challenge tasks defined within will spur further research and understanding of emoji within the multimedia community.
\ifCLASSOPTIONcompsoc
  \section*{Acknowledgments}
\else
  \section*{Acknowledgment}
\fi

Funding for this research was provided by the STW Story project.

\ifCLASSOPTIONcaptionsoff
  \newpage
\fi



\bibliographystyle{ieee}
\bibliography{papers}
%
%

%
%
%




\end{document}